%% file: main.tex
\documentclass[journal]{IEEEtran}

\usepackage{graphicx}
\usepackage{amsmath, amsfonts, amssymb}
\usepackage{booktabs}
\usepackage{multirow}
\usepackage{algorithm}
\usepackage{algorithmic}
\usepackage{hyperref}
\usepackage{color}
\RequirePackage[numbers,sort&compress]{natbib}
\usepackage{xcolor}
\usepackage[table]{xcolor}
\usepackage{makecell}
\usepackage{tikz}
\usepackage{alphalph}
\usepackage{pifont}
\usepackage{natbib}
\usepackage{xspace}
\usepackage{cleveref}
\usepackage{graphicx}
\usepackage{calc}

\definecolor{block-gray}{gray}{0.96}
\definecolor{zshot-blue}{RGB}{218,232,252}
\definecolor{ft-red}{RGB}{248,206,204}
\definecolor{lots-ca}{RGB}{203, 195, 227}
\definecolor{method-green}{RGB}{213,232,212}
\newcommand{\inlineColorbox}[2]{\begingroup\setlength{\fboxsep}{1pt}\colorbox{#1}{\hspace*{2pt}\vphantom{Ay}#2\hspace*{2pt}}\endgroup}

\newcommand{\eg}{\textit{e.g.}\xspace}
\newcommand{\ie}{\textit{i.e.}\xspace}
\def\methodshort{LOTS\xspace}
\def\dataset{Sketchy\xspace}

\include{preamble}
\usepackage{tikz}

\begin{document}

\title{Multi-Level Conditioning by Pairing Localized \\ Text and Sketch for Fashion Image Generation}
\author{authors.}

\author{
Ziyue Liu\orcidlink{0009-0004-2793-3326},
Davide Talon\orcidlink{0009-0003-6029-1532},
Federico Girella\orcidlink{0009-0001-6400-8859}, 
Zanxi Ruan\orcidlink{0000-0002-7756-8249}, \\
Mattia  Mondo\orcidlink{0009-0000-0870-1707}, 
Loris Bazzani\orcidlink{0009-0003-1970-1085}, 
Yiming Wang\orcidlink{0000-0002-5932-4371}, 
Marco Cristani\orcidlink{0000-0002-0523-6042}, 
\IEEEmembership{Member, IEEE} 
\thanks{Ziyue Liu is with the University of Verona, 37129 Verona, Italy, and also with the Polytechnic Institute of Turin, 10129 Turin, Italy.} 
\thanks{Davide Talon is with the Fondazione Bruno Kessler, 38123 Povo, Italy.}
\thanks{Federico Girella is with the University of Verona, 37129 Verona, Italy.}
\thanks{Zanxi Ruan is with the University of Verona, 37129 Verona, Italy.}
\thanks{Mattia Mondo is with the University of Verona, 37129 Verona, Italy.}
\thanks{Loris Bazzani is with the University of Verona, 37129 Verona, Italy.}
\thanks{Yiming Wang is with the Fondazione Bruno Kessler, 38123 Povo, Italy.}
\thanks{Marco Cristani is with the University of Verona, 37129 Verona, Italy, and also with the Reykjavik University, 102 Reykjavik, Iceland.}
}

{}

\maketitle

\input{sections/abstract}

\input{sections/introduction}
\input{sections/related}
\input{sections/method}

\input{sections/dataset}

\input{sections/experiments}

\input{sections/conclusion}
\input{sections/ack}

\bibliographystyle{IEEEtran}
\bibliography{references}


\newpage

\begin{IEEEbiography}[{\includegraphics[width=1in,height=1.25in,clip,keepaspectratio]{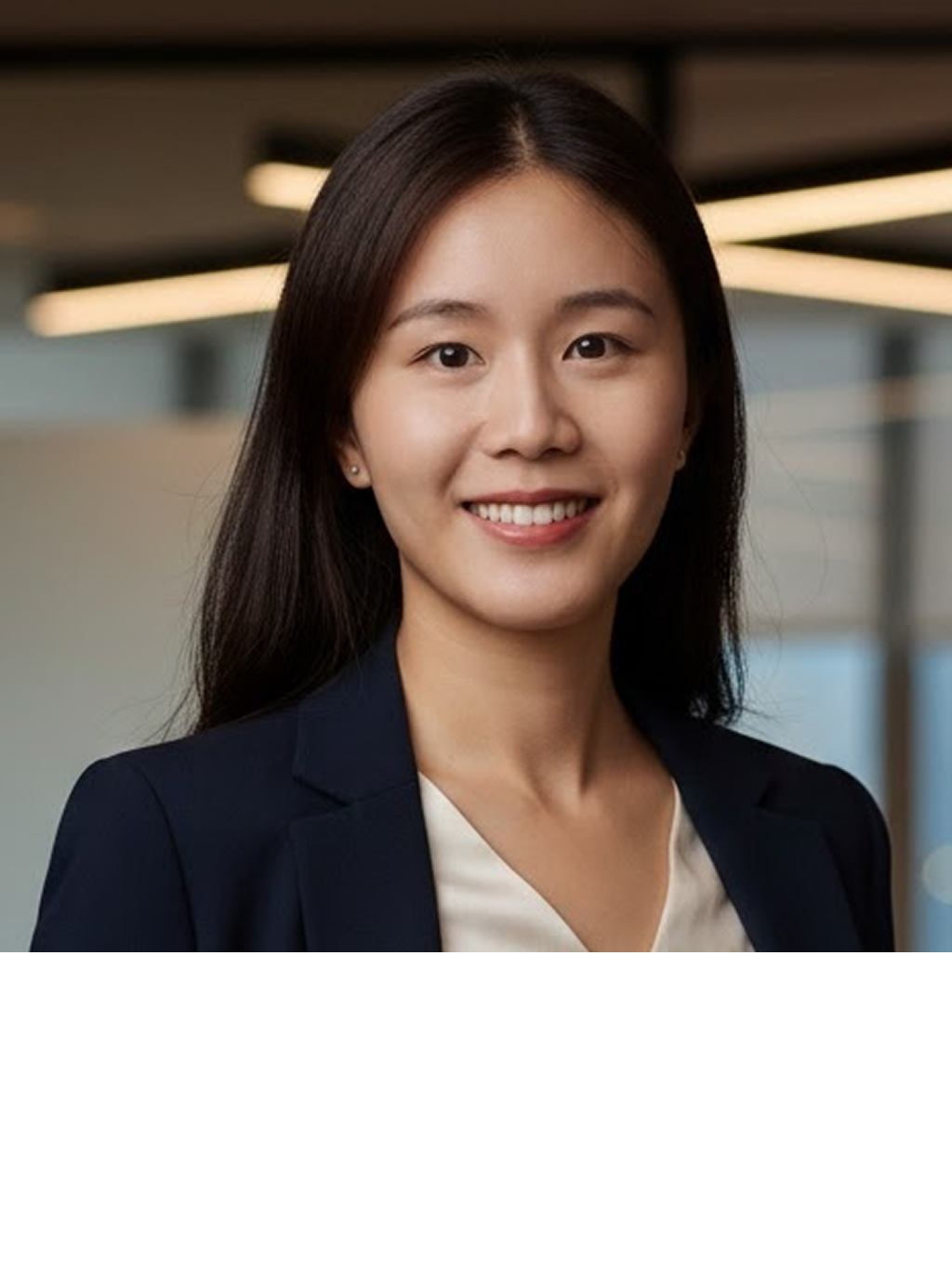}}]{Ziyue Liu}
is a student of the National PhD Programme in Artificial Intelligence at University of Verona and Polytechnic Institute of Turin, in Italy. Her research interests span the broad fields of machine learning and neural networks, with a particular focus on generative AI, multi-modal understanding and continual learning.
\end{IEEEbiography}

\begin{IEEEbiography}[{\includegraphics[width=1in,height=1.25in,clip,keepaspectratio]{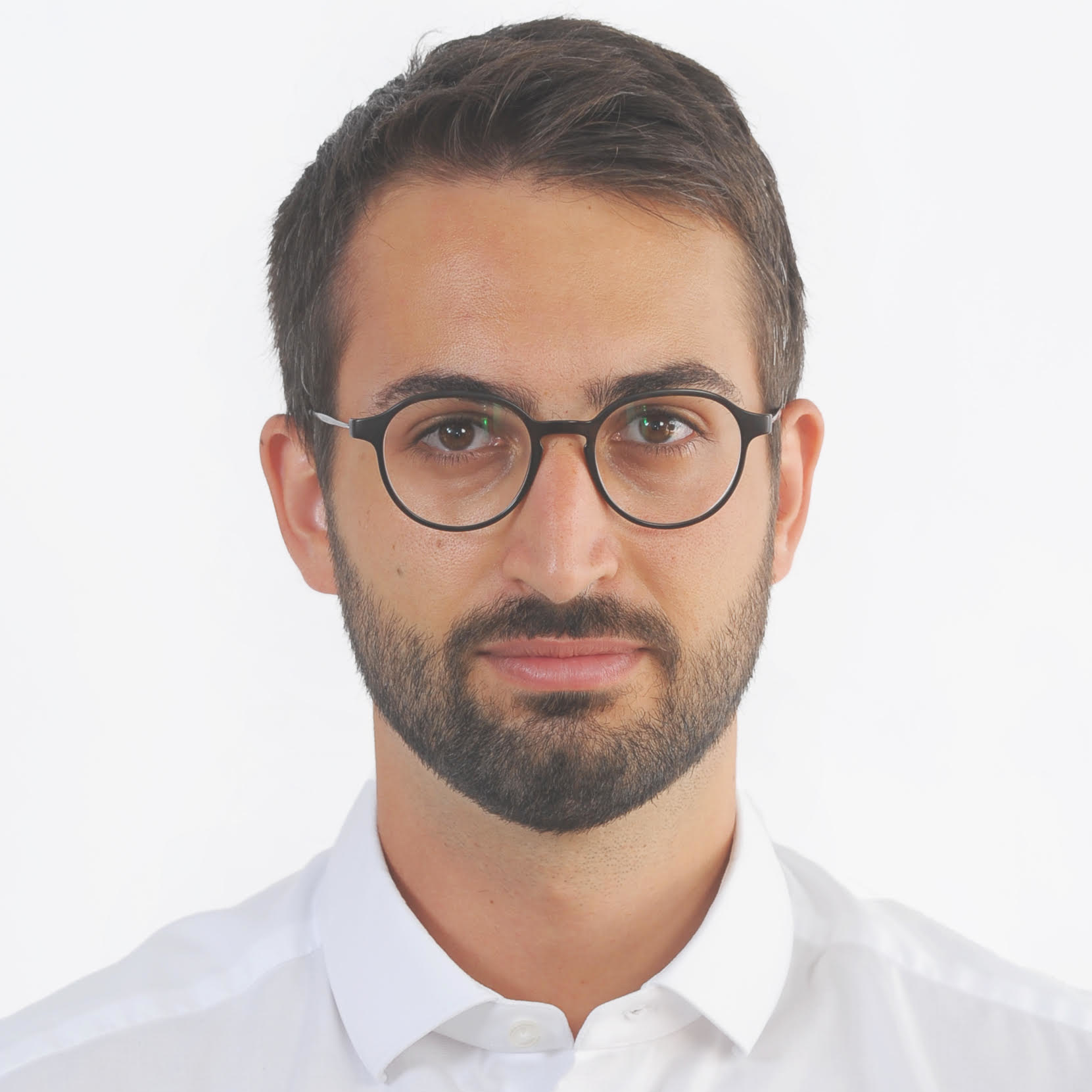}}]{Davide Talon}
is a researcher in the Deep Visual Learning (DVL) unit at Fondazione Bruno Kessler (FBK). Davide obtained his PhD in Electronics and Telecommunication Engineering from University of Genova (IT) in 2024. His research interests lie at the intersection of multimodal and representation learning. As an active member of the scientific community, Davide  serves as a reviewer for top-tier conferences and is a member of the CVF and CVPL.
\end{IEEEbiography}

\begin{IEEEbiography}[{\includegraphics[width=1in,height=1.25in,clip,keepaspectratio]{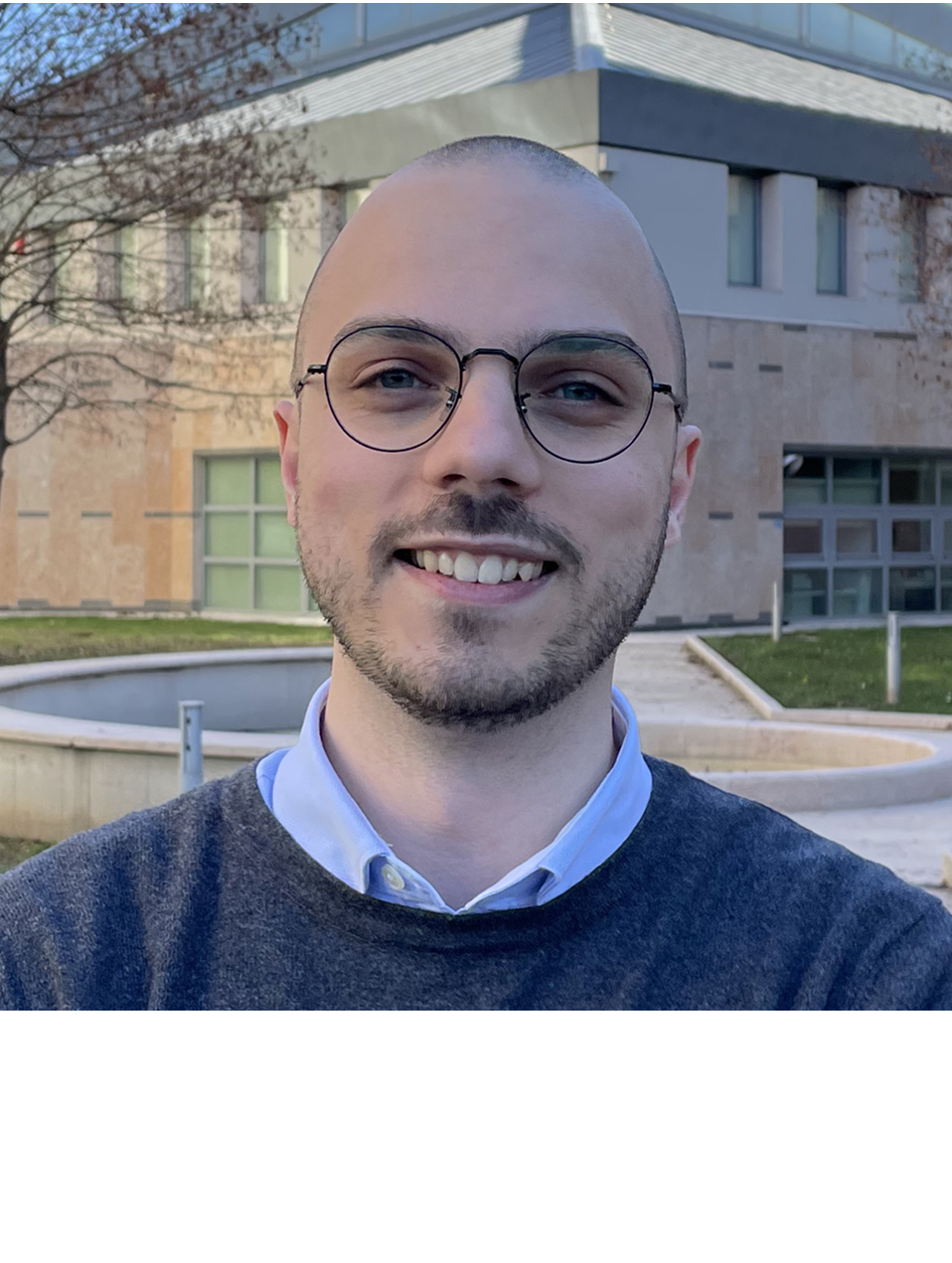}}]{Federico Girella}
received the master’s degree in computer science and engineering from the University of Verona, Verona, Italy, in 2022, where he is currently pursuing the Ph.D. degree in artificial intelligence. His main research interest includes vision and language AI for image generation and retrieval.
\end{IEEEbiography}

\begin{IEEEbiography}[{\includegraphics[width=1in,height=1.25in,clip,keepaspectratio]{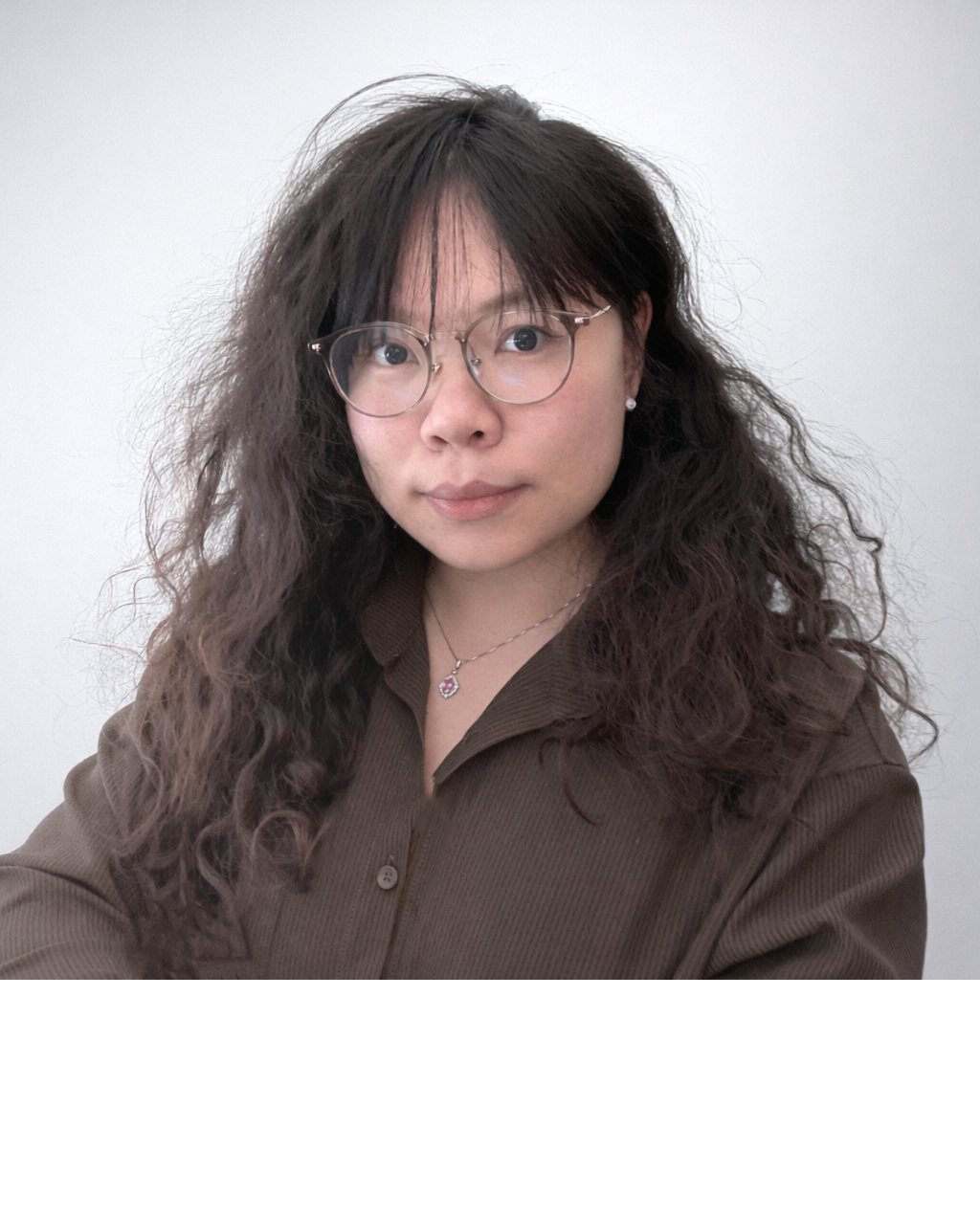}}]{Zanxi Ruan}
is a PhD student at the University of Verona, Italy, affiliated with the IntelliGO Lab in the Department of Innovation in Medical Engineering. She started her PhD in October 2024. Her research interest includes Vision-language alignment with a particular interests in multimodal learning, vision-language alignment.
\end{IEEEbiography}

\begin{IEEEbiography}[{\includegraphics[width=1in,height=1.25in,clip,keepaspectratio]{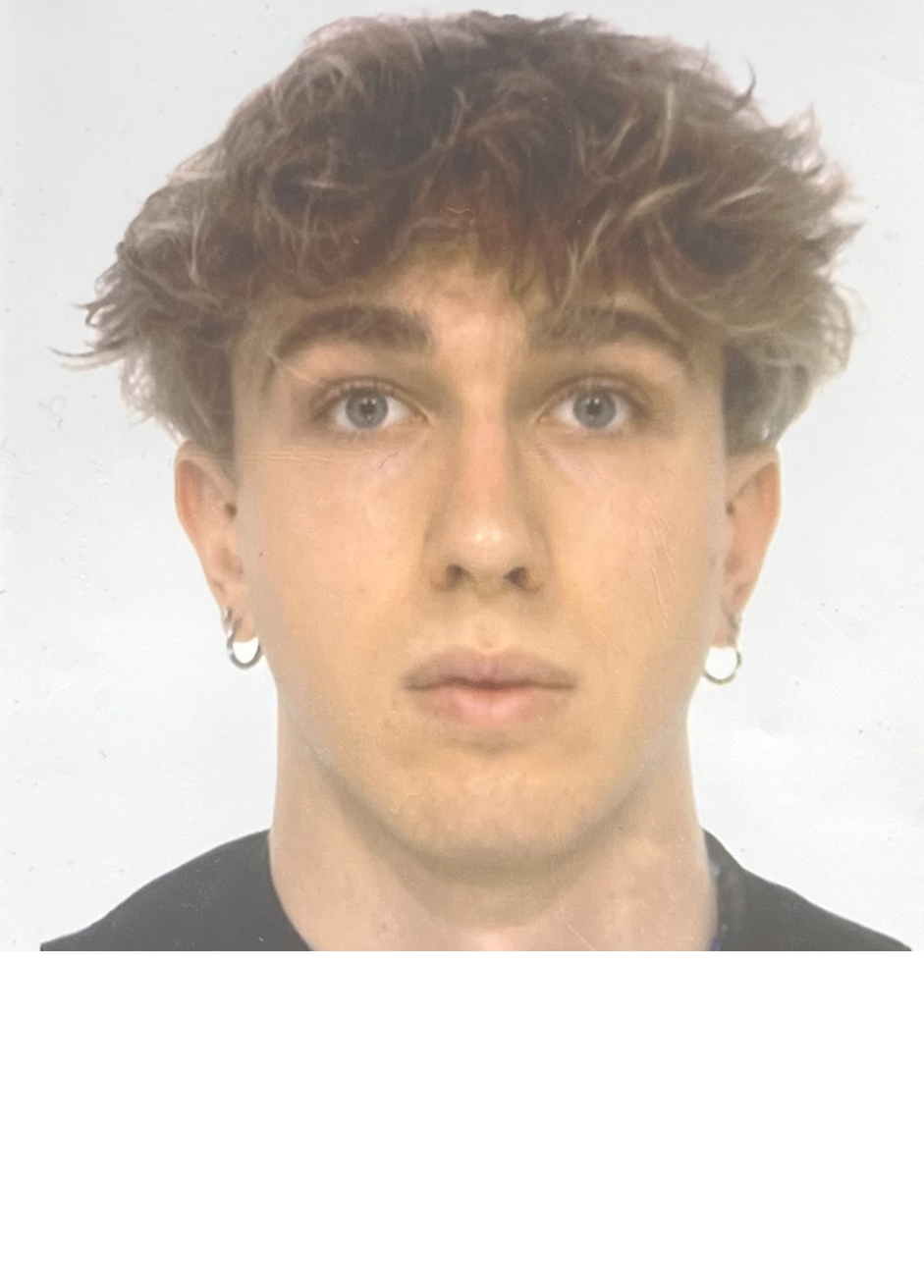}}]{Mattia Mondo}
is a Master's student in Artificial Intelligence at the University of Verona. He earned his Bachelor's degree in Computer Science in 2025. His main interests include Computer Vision, Large Language Models (LLMs), Machine Learning, and web application development.
\end{IEEEbiography}

\begin{IEEEbiography}
[{\includegraphics[width=1in,height=1.25in,clip,keepaspectratio]{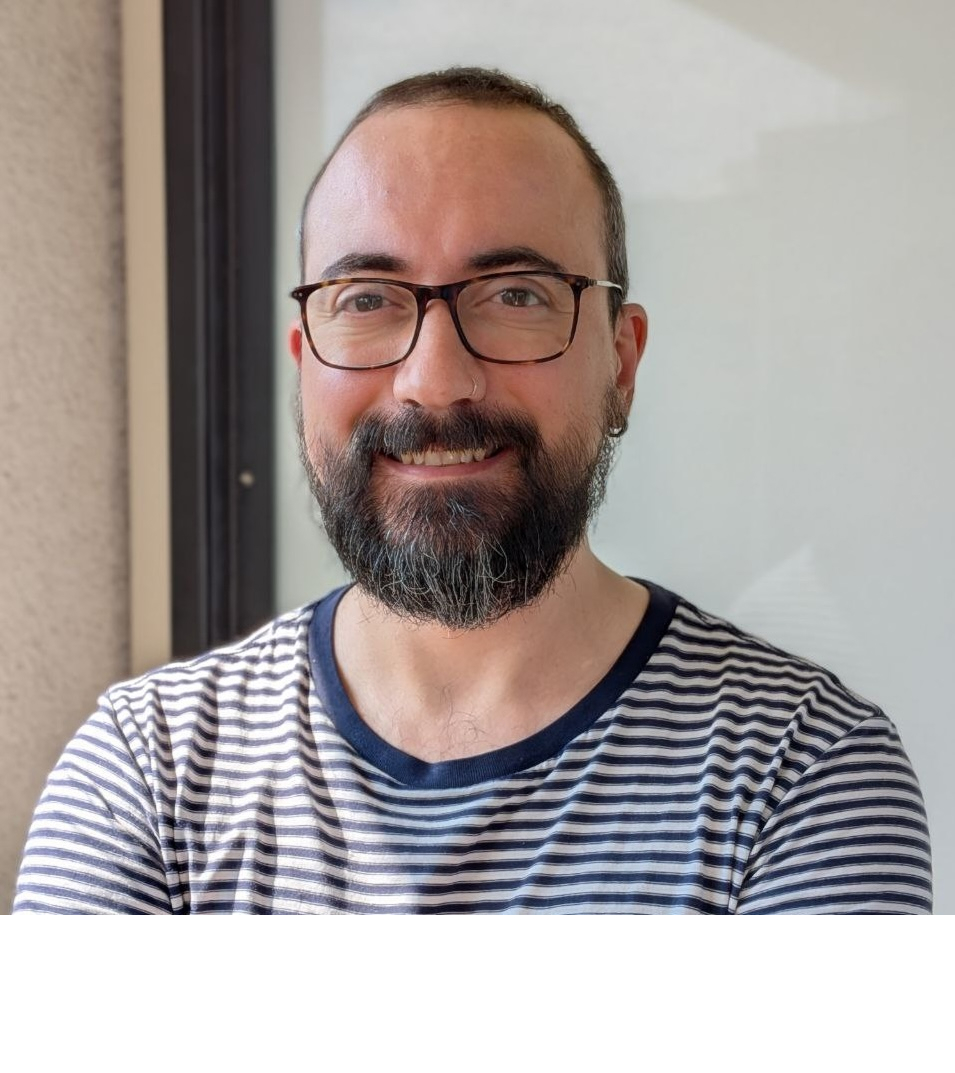}}]{Loris Bazzani} 
is a research leader with over 15 years of experience in AI, currently an adjunct professor at University at Verona. Previously, he was a Principal Scientist at Amazon, leading research and product efforts on video understanding, vision-language representation, large multimodal models, and diffusion models, powering novel features in live sports highlights, virtual try-on, interactive fashion recommendations, and shopping assistants. 
Loris obtained his Ph.D. from University of Verona and held postdoc positions at Dartmouth College and at the Italian Institute of Technology (IIT). 
\end{IEEEbiography}

\vspace{-15px}
\begin{IEEEbiography}[{\includegraphics[width=1in,height=1.25in,clip,keepaspectratio]{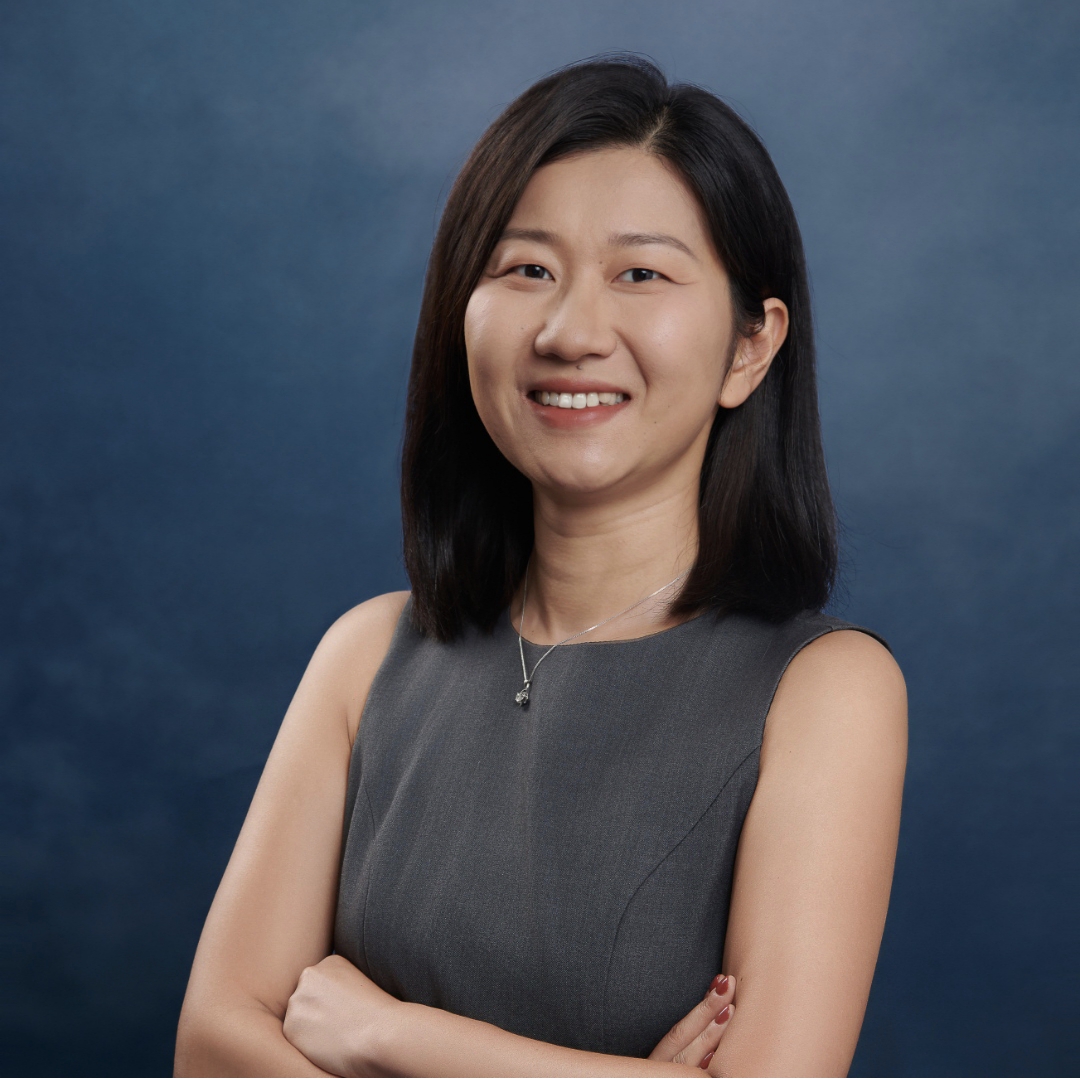}}]{Yiming Wang}
is a researcher in the Deep Visual Learning (DVL) unit at Fondazione Bruno Kessler (FBK). Yiming obtained her PhD in Electronic Engineering from Queen Mary University of London (UK) in 2018. She works on 
topics related to vision-language scene understanding and robotic perception. She is actively contributing to field as area chairs and serving as reviewer for top-tier conferences and journals in both the Computer Vision and Robotics domains. She is a member of ELLIS.
\end{IEEEbiography}

\vspace*{-5pt}

\begin{IEEEbiography}[{\includegraphics[width=1in,height=1.25in,clip,keepaspectratio]{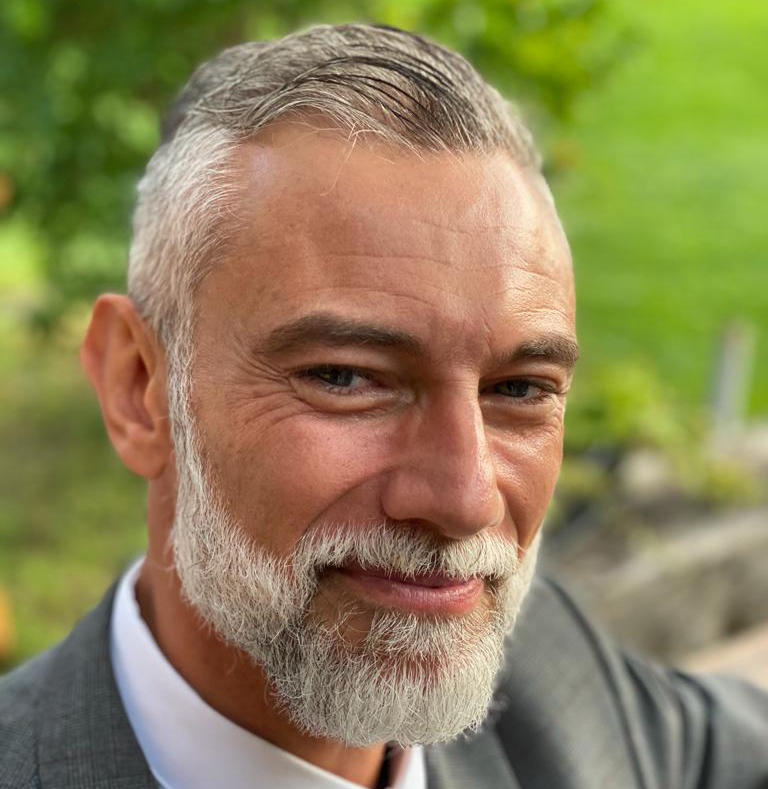}}]{Marco Cristani} is Full Professor (Professore Ordinario) at the Dept. of Engineering for Innovation Medicine, University of Verona,
Associate Member at the National Research Council (CNR), External Collaborator at the Italian Institute of Technology (IIT). His main research interests are in statistical pattern recognition and computer vision, mainly in deep learning and generative modeling, with application to social signal processing and fashion modeling. On these topics he has published more than 200 papers. He has organised 11 international workshops.
He is or has been the Principal Investigator of several national and international projects, including PRIN and H2020 projects. He is an IAPR fellow and a member of IEEE.
\end{IEEEbiography}

\vspace*{262pt}
\end{document}

%% file: preamble.tex
\newcommand{\range}[2]{#1,.., #2}

\usepackage{orcidlink}

\newcommand{\iconG}{\raisebox{-0.3em}{\includegraphics[height=1.2em]{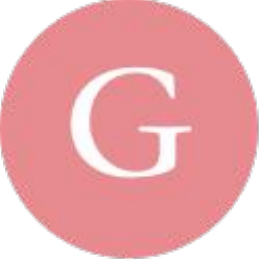}}}
\newcommand{\iconL}{\raisebox{-0.3em}{\includegraphics[height=1.2em]{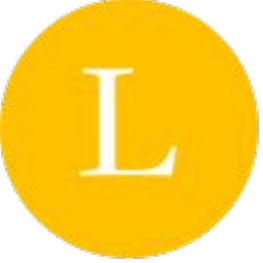}}}

%% file: sections/abstract.tex
\begin{abstract}
Sketches offer designers a concise yet expressive medium for early-stage fashion ideation by specifying structure, silhouette, and spatial relationships, while textual descriptions complement sketches to convey material, color, and stylistic details. Effectively combining textual and visual modalities requires adherence to the sketch visual structure when leveraging the guidance of localized attributes from text. 
We present \emph{LOcalized Text and Sketch with multi-level guidance} (LOTS), a framework that enhances fashion image generation by combining global sketch guidance with multiple localized sketch–text pairs. LOTS employs a Multi-level Conditioning Stage to independently encode local features within a shared latent space while maintaining global structural coordination. Then, the Diffusion Pair Guidance stage integrates both local and global conditioning via attention-based guidance within the diffusion model’s multi-step denoising process.
To validate our method, we develop \dataset, the first fashion dataset where multiple text-sketch pairs are provided per image. \dataset provides high-quality, clean sketches with a professional look and consistent structure. To assess robustness beyond this setting, we also include an “in the wild” split with non-expert sketches, featuring higher variability and imperfections.
Experiments demonstrate that our method strengthens global structural adherence while leveraging richer localized semantic guidance, achieving improvement over state-of-the-art. The dataset, platform, and code are publicly available at \url{https://intelligolabs.github.io/lots/}.
\end{abstract}


%% file: sections/introduction.tex
\section{Introduction}
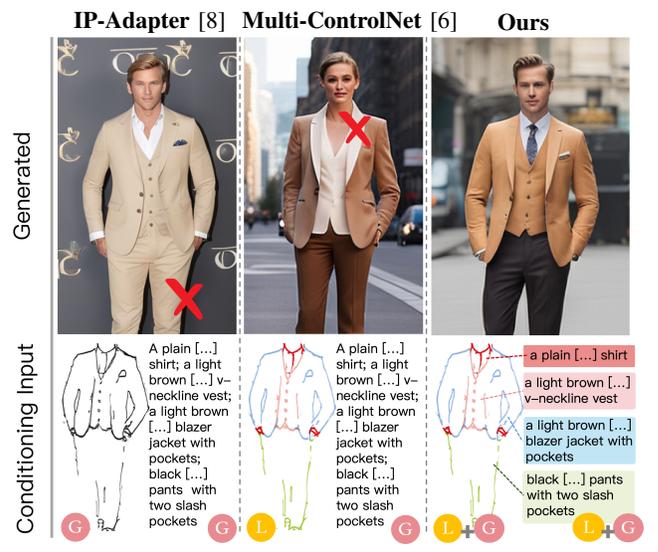
\begin{figure}
    \centering
    \begin{tikzpicture}
    \node[anchor=south west, inner sep=0] (image) at (0,0) {\includegraphics[width=\linewidth]{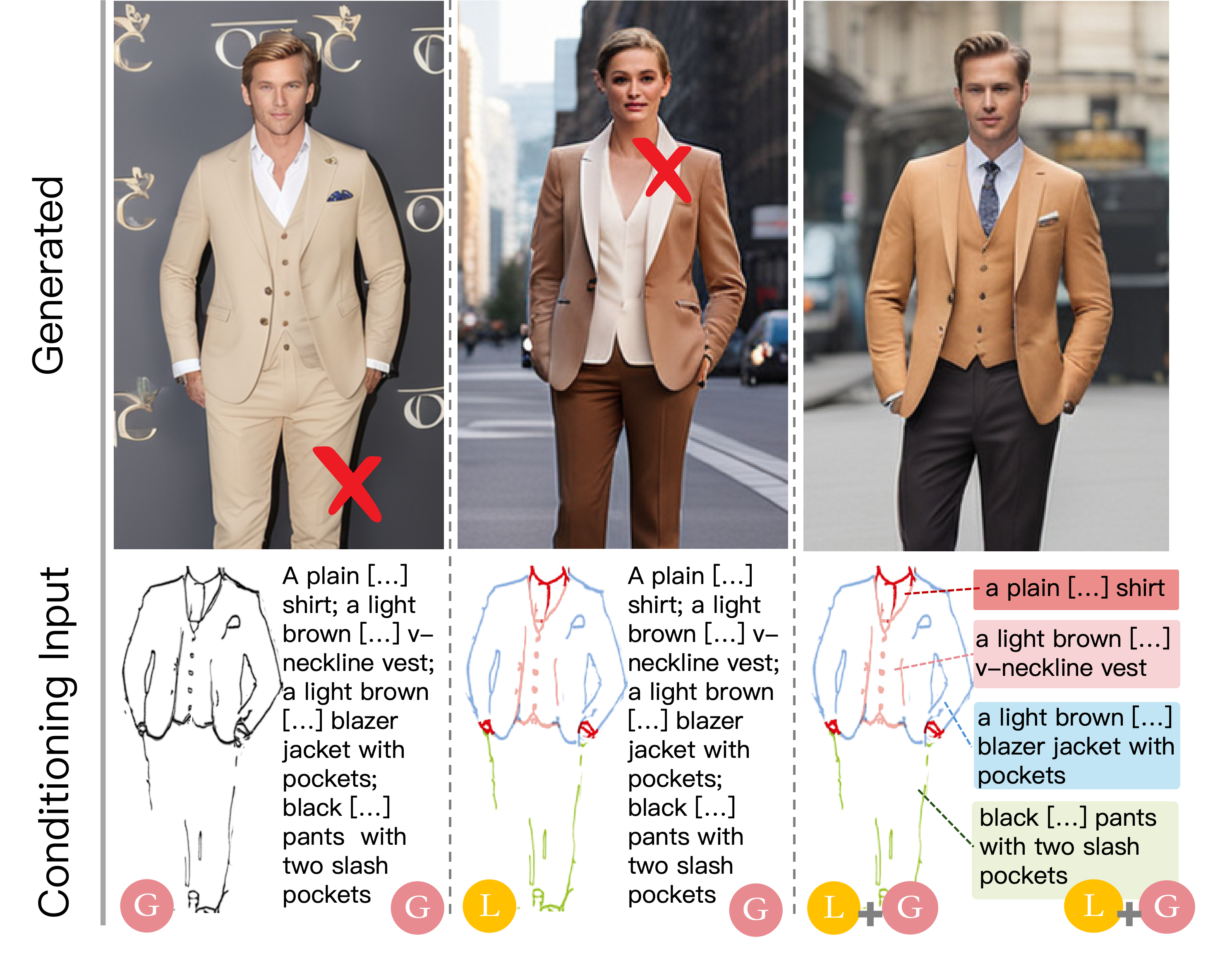}};
        \begin{scope}[x={(image.south east)}, y={(image.north west)}]
            \node at (0.23, 1.03) {\small \textbf{IP-Adapter}~\citep{ye2023ip}};
            \node at (0.53, 1.03) {\small \textbf{Multi-ControlNet}~\citep{zhang2023adding}};
            \node at (0.79, 1.03) {\small \textbf{Ours}};
        \end{scope}
    \end{tikzpicture}
    \caption{LOTS enables automation of the fashion design process at a new level of detail. The figure illustrates a design scenario where sketches are complemented by natural language descriptions to characterize garment material, style, and structure. LOTS represents a paradigm shift in design methodologies, advancing from global (\iconG) text with global sketch (IP-Adapter~\citep{ye2023ip}) and global text with localized (\iconL) sketches (Multi-ControlNet~\citep{zhang2023adding}). Our approach adds localized sketch-text specifications (the coloured boxes), enabling fine-grained control over the layout and attributes of multiple garment items. All textual descriptions are shown in a contracted form for readability, see text.}
    \label{fig:teaser}
\end{figure}

\IEEEPARstart{S}
ketching is a fundamental medium in early-stage fashion design, providing an expressive visual representation of proportions, silhouette, spatial layout, and structural details~\cite{shevchuk2025sketching}. Textual descriptions often complement visual sketches by conveying semantic attributes that are difficult or impossible to express visually, such as material and stylistic pattern~\cite{lau2009sketching,guo2023ai}. 
For example, as illustrated in Fig.~\ref{fig:teaser}, a designer can sketch the silhouette of a vest and provide a natural language description such as ‘‘\textit{a light brown, single-breasted, tight-fitting vest with a V-neckline, a normal waist, an above-the-hip length, and a symmetrical design}’’. 
Similarly, the designer can describe the details of the shirt and the pants. 
Recently, generative models have begun to automate parts of this design process, enabling the synthesis of realistic fashion images directly from sketches and textual descriptions~\cite{rombach2022high,podell2023sdxl,zhang2023adding,mou2024t2i,ye2023ip,sun2024anycontrol}.

In practice, a complete fashion design typically comprises multiple garments. Accordingly, designers often collect several sketch-text pairs, each specifying a \textit{localized part} of the overall design (\eg, an individual garment), thereby enabling fine-grained localized control over the design process. In this work, we aim to support the concretization of such design ideas into graphical outputs by leveraging \textit{localized sketch-text conditions}. We formulate this setting as a conditional image generation task in which the conditioning consists of a set of localized sketch-text pairs. To emphasize the presence of multiple localized conditions, we refer to this problem as \emph{multi-localized} conditional image generation.


In addressing this problem,
state-of-the-art methods fall short. 
Recent diffusion adapters for sketch-to-image generation~\cite{mou2024t2i, ye2023ip, sun2024anycontrol, zhang2023adding} allow for multi-spatial conditions but underperform when providing fine-grained textual information, such as neckline type and pattern style, as demonstrated in Fig.~\ref{fig:teaser}.
We argue that this limitation stems from the use of a single global description to inject textual conditions: all relevant details about different parts of the outfit are considered in a monolithic fashion, leading to incorrect localization of attributes to parts~\citep{mou2024t2i, ye2023ip, sun2024anycontrol, zhang2023adding}. 
We refer to this problem as \emph{attribute confusion}, following~\cite{liu2025evaluating}, where properties of one item are incorrectly generated for another, \eg, ‘‘\textit{a light brown blazer jacket}’’ and ‘‘\textit{black pants}’’ result in the light brown color appearing on the pants (Fig.~\ref{fig:teaser}, first two columns).


From another perspective, jointly leveraging multiple local sketches and text descriptions coherently is challenging. When several sketch text pairs define different garment parts, the model must associate each description with the correct sketch region while preserving the overall outfit structure from the sketch. Failure can result in misaligned silhouettes, attribute leakage, or loss of the original sketch structure.

In this paper, we address the multi-localized conditional image generation problem by introducing LOcalized Text and Sketch with multi-level guidance (\methodshort), the first approach explicitly designed for multi-localized sketch-text semantic conditioning. \methodshort takes as input a set of localized sketches paired with textual descriptions. Taken together, the local sketches form a global sketch, which is accompanied by a global context description specifying overall stylistic attributes and background characteristics (e.g. ‘‘a male subject in a metropolitan scenario’’).   
\methodshort operates through a \textit{Multi-level Conditioning} strategy that treats local pairs and global structure as distinct yet complementary signals. At the local level, we introduce a \textit{Modularized Pair-Centric Representation}: sketch-text pairs are first embedded via modality-specific encoders and then fused by the \textit{Pair-former} using learnable tokens to produce spatially grounded representations. Simultaneously, at the global level, a \textit{Global Conditioning} module captures the overall sketch structure to provide high-level context via cross-attention. In a second stage, through \textit{Diffusion Pair Guidance}, these dual-level representations are injected into the diffusion process across multiple denoising steps. This approach ensures coherent generation that preserves global structural integrity while preventing the explicit merging of local pairs, effectively reducing attribute confusion.


For training and evaluation, we introduce \dataset{}, the first dataset specifically designed for localized sketch-to-image generation. \dataset{} is built upon Fashionpedia~\citep{jia2020fashionpedia}, which we restructure and enrich to support localized sketch-text conditioning. In \dataset{}, garments within the same outfit are treated as multiple conditioning inputs, each paired with a fine-grained textual description and a corresponding sketch (47K outfits, 104K localized pairs). The \dataset{} annotations are generated automatically to resemble professional fashion sketches, modeling the structure and proportions typical of trained designers. We also consider sketches produced by non-professional users, introducing a dedicated split composed of casual, in-the-wild drawings collected from a general audience (141 outfit sketches, 2.8k localized pairs). This split enables evaluation under realistic, imperfect inputs, assessing robustness to variability and noise in sketching.

We evaluate \methodshort against state-of-the-art baselines on \dataset, assessing global image quality, sketch adherence, and localized semantic alignment. Results show that \methodshort consistently achieves the best overall trade off, with top performance on global and garment level alignment and strong robustness to casual sketches under domain shift, confirmed by ablations and human studies.  
\vspace{5pt}

\noindent{\textbf{Our contributions}} are four-fold:
\begin{itemize}
\item We establish a new formulation for sketch-text image generation that enables \emph{fine-grained}, \emph{local-level} control by leveraging multiple localized sketches-text pairs.

\item We propose \methodshort{}, a novel multi-level conditioning framework that processes sketch-text pairs independently and integrates them during denoising, effectively mitigating attribute leakage while preserving global 
adherence.

\item We introduce \dataset{}, a large-scale fashion dataset designed for multi-localized sketch-text conditioning. Casual sketches in its in-the-wild split enable rigorous evaluation in a general audience scenario.

\item Extensive experiments demonstrate that \methodshort{} sets a new state-of-the-art in localized sketch-text generation, achieving superior garment-level semantic alignment, strong sketch adherence, and robust generalization to casual sketches, as confirmed by quantitative metrics and human studies.
\end{itemize}
This paper presents a substantial extension of the conference version~\cite{girella2025lots}. Beyond the original formulation, we introduce a multi-level localized sketch–text conditioning strategy that explicitly reinforces global structural guidance while preserving fine-grained, localized garment-level semantic control. We further extend the \dataset{} dataset with instance-level color annotations, additional garment categories, and a new split containing casual sketches collected via mouse and stylus, supported by a dedicated interactive platform for sketch collection. Finally, we investigate generalization to casual sketches and extend the validation adopting an additional evaluation metric, Localized-VQAScore, to quantify garment-level semantic alignment and attribute localization based on Visual Question Answering. 

%% file: sections/related.tex
\section{Related Works}
\label{sec:related}
This section surveys prior work on text-to-image and sketch-to-image generation, as well as diffusion-based methods for controllable fashion synthesis. Finally, we review the available datasets in the literature for fashion image generation.

\noindent\textbf{Text-to-Image Generation.}
Recent progress in Text-to-Image (T2I) generation has been largely driven by diffusion models~\cite{ho2020denoising,ho2022classifier,song2020denoising}, which generate high-quality images from textual prompts~\cite{nichol2021glide,rombach2022high,ramesh2022hierarchical,saharia2022photorealistic}. These models operate through a forward process that incrementally adds noise to images and a learned reverse process that reconstructs coherent outputs through denoising.
Early works such as 
GLIDE~\cite{nichol2021glide} adopt classifier-free guidance to improve sample quality, while DALLE-2~\cite{ramesh2022hierarchical} introduces a two-stage pipeline to generate images from CLIP embeddings.
Similarly, Imagen~\cite{saharia2022photorealistic} integrates large-scale language models to improve realism and semantic alignment. 
More recently, Stable Diffusion (SD)~\cite{rombach2022high} refines conditioning via cross-attention mechanisms while balancing computational efficiency and detail preservation through latent-space diffusion.
Building on SD–like architectures, we extend control beyond text by introducing multi-level conditioning that jointly leverages complementary text and sketch modalities.

\noindent\textbf{Sketch-to-Image Generation.}
Early sketch-to-image methods were predominantly based on GAN architectures~\citep{isola2017image, lu2018image, ghosh2019interactive, koley2023picture, richardson2021encoding}, while more recent approaches have shifted toward large-scale pre-trained diffusion models~\citep{wang2022pretraining, voynov2023sketch, mengsdedit}. 
Among these, PITI~\cite{wang2022pretraining} maps sketch inputs into the semantic latent space of diffusion models, whereas SDEdit~\citep{mengsdedit} performs generation by injecting noise into sketches and iteratively denoising them toward realistic images. 
LGP~\citep{voynov2023sketch} further improves alignment by explicitly maintaining spatial correspondence between sketch guidance and intermediate noisy features.
More recent work has instead investigated alternative design choices for sketch-conditioned diffusion, including explicit spatial control~\citep{zhang2023adding, mou2024t2i, ye2023ip}, varying levels of sketch abstraction~\citep{navard2024knobgen, koley2024s}, and the use of professional or line-art sketches~\citep{wang2024lineart}. 
In contrast to existing approaches~\citep{zhang2023adding, mou2024t2i, ye2023ip, navard2024knobgen, koley2024s, wang2024lineart, wang2022pretraining, mengsdedit, voynov2023sketch}, which rely on global sketch conditioning, our method enables localized sketch-based control.

\noindent\textbf{Controllable Diffusion-based Generation.}
While textual prompts enable high-quality image generation in T2I models, they often fall short in providing fine-grained control. 
To improve controllability, a broad line of work augments diffusion models with additional conditioning elements~\cite{zhang2023adding, huang2023t2i, ye2023ip, li2023gligen, sun2024anycontrol, zhao2024uni}, including bounding boxes~\citep{li2023gligen}, spatial blobs~\citep{nie2024compositional}, and segmentation masks~\citep{kim2023dense, goel2024pair}.
GLIGEN~\citep{li2023gligen} conditions the model with bounding box coordinates to localize textual concepts, but does not allow for paired sketch-text localization.
ControlNet~\citep{zhang2023adding} 
modulates a frozen diffusion backbone through zero-convolution layers, while subsequent works extend this paradigm to multimodal~\cite{hu2023cocktail} or unified control settings~\cite{zhao2024uni}, both rely on fixed-length input channels. 
AnyControl~\cite{sun2024anycontrol} enables flexible multi-condition guidance but requires training a copy of the diffusion model. 
Adapter-based approaches, such as T2I~\citep{mou2024t2i} and IP~\citep{ye2023ip} adapters fuse available multi-conditions and later adopt residual feature maps or cross-attention for diffusion steering. 
However, these methods rely on global textual prompts and remain constrained by the 77-token limit of text encoders. 
In contrast, we couple localized textual descriptions with their corresponding sketches to allow for fine-grained generation. Our adapter design enables a pre-trained T2I diffusion model to condition on a variable number of sketch-text pairs while remaining lightweight to train.

\noindent\textbf{Fashion Image Generation.}
Recent advances for fashion generation build on multimodal conditioning, significantly improving visual quality and semantic alignment to input conditions~\cite{zhang2024garmentaligner,guo2025higarment,pathak2024controllable}.
GarmentAligner~\citep{zhang2024garmentaligner} improves generation text-consistency leveraging a retrieval augmented pipeline. 
GenWear~\citep{pathak2024controllable} encodes spatial priors from the global design sketch and injects it into a frozen diffusion backbone for structural fidelity.
HiGarment~\citep{guo2025higarment} employs an attention mechanism to align sketch with realistic textures, synthesizing an image under the structural constraints of flat design drawings.
While existing methods advance fine-grained single-garment synthesis, their reliance on image-level control often causes attribute confusion in multi-garment outfits, thereby failing to capture the compositionality required for complex fashion synthesis in realistic settings~\cite{liu2025evaluating,girella2025lots}.
Several fashion-specific approaches~\citep{xie2025hierafashdiff,Baldrati_2023_ICCV} apply iterative design workflows to improve controllability in generated images.
HieraFashDiff~\citep{xie2025hierafashdiff} is a recent work presenting a two-stage pipeline performing generation and iterative editing. 
Similarly, Multimodal Garment Designer~\citep{Baldrati_2023_ICCV} requires a starting image as input for the edit. 
These methods differ from \methodshort in that they perform image editing starting from an existing image, with the aim of modifying only parts of this image, whereas \methodshort generates images from scratch.
Indeed, image editing is generally regarded as a separate task that follows the initial generation stage.
In this work, we focus on the task of one-shot controllable image generation for complex multi-garment outfits.

\noindent\textbf{Datasets for Fashion Image Synthesis.}
Fashion image synthesis is supported by a diverse datasets that differ both in the type of conditioning signal (text, attributes, segmentation masks, or sketches) and in annotation granularity (single- vs multi-garment). Most existing fashion datasets focus on single-garment scenarios~\citep{liuLQWTcvpr16DeepFashion, rostamzadeh2018fashion, choi2021viton, morelli2022dress}, and therefore fail to capture the interactions and compositional dependencies that arise in full outfits composed of multiple garments.
Representative single-garment datasets include DeepFashion~\citep{liuLQWTcvpr16DeepFashion}, which provides paired images and textual descriptions but lacks sketch annotations, and FashionGen~\citep{rostamzadeh2018fashion}, which associates product images with concise, attribute-centric captions but fails to capture detailed fashion nuances. Similarly, virtual try-on (VTON) datasets~\citep{morelli2022dress, choi2021viton} operate at the single-garment level, emphasizing identity and appearance preservation of a provided item when transferred onto a person, rather than garment synthesis from multimodal inputs such as sketch–text pairs. Extensions of these datasets~\citep{baldrati2023multimodal, baldrati2024multimodal} collect automated sketches but remain limited to single-garment annotations.
In contrast, Fashionpedia~\citep{jia2020fashionpedia} and DeepFashion2~\citep{DeepFashion2} provide outfit-level annotations with multiple garments per image, including fine-grained segmentations and category labels, with Fashionpedia further annotating garment attributes. 
\dataset extends Fashionpedia’s multi-garment supervision with localized sketch annotations, collected in-the-wild for a subset of images, 
and detailed, garment-specific textual descriptions that include color and appearance attributes.  Hence, \dataset enables fine-grained, localized multimodal control for fashion image generation.

%% file: sections/method.tex
\section{Method}
\label{sec:method}
\begin{figure*}
    \centering
    \includegraphics[width=1.0\linewidth]{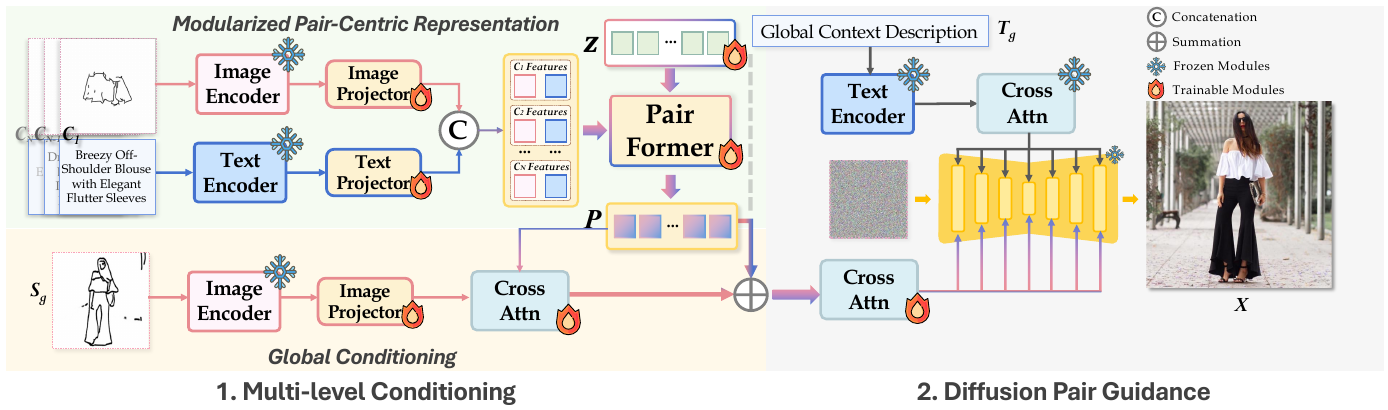} 
\caption{\methodshort{} pipeline. \textbf{1.} The first \textit{Multi-level conditioning } stage constructs a conditioning representation spanning both local and global levels. Locally, the \emph{Modularized Pair-Centric Representation} module (Sec.~\ref{sec:method:pair-centric-representation}) handles each sketch–text pair independently: modality-specific, frozen encoders first map sketches and texts into their respective embeddings, which are then fused in the Pair-Former by integrating textual semantics with the spatial structure of the corresponding sketch. In parallel, the \textit{Global Conditioning} (Sec.~\ref{sec:method:global-conditioning}) derives a global representation from the full sketch and injects it via cross-attention to promote consistency and interaction across multiple pairs. \textbf{2.} In the subsequent \emph{Diffusion Pair Guidance} stage (Sec.~\ref{sec:method:diffusion-guidance}), the multi-level embeddings are progressively incorporated into the diffusion process, together with the Global Context Description which drives the background generation and shapes the overall style. Rather than explicitly merging all pair representations upfront, conditioning is applied throughout the denoising process, enabling gradual integration and preventing the attribute leakage typically induced by early representation fusion.}

    
    \label{fig:method}
\end{figure*}
In this section, we start with the task formulation. We then present the proposed method \textbf{\methodshort} with multi-level localized text and sketch conditioning for fashion image generation.


\subsection{Problem Formulation} 
Let $\mathcal{C} = \{C_1, \dots, C_N\}$ denote $N$ localized sketch-text pairs for a given sample, where $C_i = (S_i, T_i)$ consists of the $i$-th sketch $S_i \in \{0,1\}^{H \times W}$ with $H$ and $W$ representing the height and width, respectively, and $T_i$ is the associated textual description. We assume the local sketches are spatially coherent and compose to the global sketch. In practice, the global sketch $S_g$ is the union of all local sketches: $S_g = \bigcup_{i=1}^{N} S_i$. We further allow for a global context description $T_g$ to provide the model with general appearance information, such as overall fashion style or background specification.
Multi-localized conditional image generation aims to train a generative model $\phi$ to synthesize an image $X \in \mathbb{R}^{3 \times H \times W}$ conditioning on localized sketch-text pairs $\mathcal{C}$, as well as the global sketch $S_g$, and global context description  $T_g$.
Formally:
\begin{equation}
    X = \phi(\mathcal{C}, S_g, T_g).
\end{equation}
The generated image should accurately satisfy both global and local conditioning. At the global level, it should adhere to the overall textual description while preserving the coherent structure defined by the global sketch. At the local level, sketch–text associations must be maintained: for each localized pair, the text-specified conditioning $T_i$ of the $i$-th item should be reflected in the spatial region indicated by $S_i$, without leaking to other items $S_j$, $i, j=\range{1}{N}, j \neq i$. 

\smallskip
\noindent\textbf{Method overview.} \methodshort performs multi-localized conditional image synthesis through a two-stage pipeline, illustrated in Fig.~\ref{fig:method}.
First, the \textit{Multi-level Conditioning} stage embeds the conditioning input into a representation that jointly models local and global information. At the local level, the \textit{Modular Pair-Centric Representation} (Sec.~\ref{sec:method:pair-centric-representation}) processes localized sketches and associated textual descriptions independently using frozen, modality-specific encoders. These representations are then fused into a multimodal embedding via the \textit{Pair-Former} module, which explicitly isolates each sketch–text pair to enable independent pair modeling while preventing cross-pair interference.
However, localized conditioning alone can struggle to capture global coherence across multiple items. Therefore, we introduce a multi-level guidance scheme that extends our previous work~\cite{girella2025lots} which primarily focused on local representation: at global level, the novel \textit{Global Conditioning} branch (Sec.~\ref{sec:method:global-conditioning}) encodes the global sketch to reinforce structural consistency and promote coherent items composition.
After the multi-level conditioning, the \textit{Diffusion Pair Guidance} module (Sec.~\ref{sec:method:diffusion-guidance}) mitigates \textit{attribute confusion} arising from multiple conditioning signals by injecting both localized and global cues into the iterative denoising process via attention-based conditioning, enabling their progressive and coherent integration throughout generation.

\subsection{Multi-level Conditioning Stage} 
\label{sec:method:pair-centric-representation}
\noindent\textbf{Local Level: Modularized Pair-Centric Representation.}
To ensure that semantic information from each local pair $C_i$ does not leak to unrelated regions, we propose to process each pair in a modularized fashion where pairs do not influence each other. Each local sketch-text pair $C_i = (S_i, T_i)$ is encoded independently using modality-specific encoders, and projected into a shared latent space:
\begin{equation}
h_i^T = W_T f_T(T_i), \quad 
h_i^S = W_S f_S(S_i)
\end{equation}
where $f_T$ and $f_S$ denote pre-trained text and sketch encoders, and $W_T$ and $W_S$ are learnable projection matrices mapping the encoded features into a shared latent space. Latent representations $h_i^T$ and $h_i^S$ are generated for each pair $i = 1,\dots,N$. 

These local representations are fused into a multimodal embedding via the \textit{Pair-Former} module. Inspired by recent advancements in multimodal representations~\citep{li2023blip}, we start from modality-specific representations and adopt self-attention to compress the sparse sketch embeddings $h_i^S$ into a fixed-size representation, while integrating its associated textual information $h_i^T$. Specifically, let $z \in \mathbb{R}^{k \times d}$ be a set of $k$ learnable tokens prepended to the concatenated sketch and text embeddings. The pair tokens are obtained by applying self-attention to $[z; h_i^S; h_i^T]$, and the first $k$ tokens are retained as the output representing the localized pair. While the learnable tokens $z$ are shared across pairs, each of the $N$ input pairs are processed independently. Formally, the pair tokens for the $i$-th sketch-text pair are computed as:
\begin{equation}
    p_i = \text{SelfAttn}([z; h_i^S; h_i^T])[1:k],
\end{equation}
where $p_i \in \mathbb{R}^{k \times d}$ are the first $k$ tokens of the output associated with $z$, effectively pooling the fused pair information.

\smallskip
\noindent\textbf{Global Level: Global Conditioning.}
\label{sec:method:global-conditioning}
While localized sketch–text conditioning enables fine-grained control over individual garments, relying exclusively on local cues can hinder global coherence across the generated outfit. In particular, independently conditioned regions may lead to inconsistencies in overall pose, or outfit composition, as no mechanism explicitly enforces the global coordination among items. To address this limitation, we introduce a novel global conditioning branch that extends our previous work~\citep{girella2025lots}.
Global conditioning complements localized sketch–text pairs with global structural guidance.
To this end, the global sketch $S_g$ is encoded and projected to the same latent space as the pair tokens:
\begin{equation}
    h_g^S = W_g f_S(S_g)
\end{equation}
where $f_S$ is the pre-trained sketch encoder and $W_g$ is a learnable projection matrix.
In a second step, we introduce a cross-attention mechanism that fuses the global sketch representation $h_g^S$ with localized pair representations, enabling the model to capture high-level structural coherence while maintaining pair-specific semantics without interference. Let $P$ be the concatenation of pair tokens $P = [p_1; \dots; p_N] \in \mathbb{R}^{(N \cdot k) \times d}$.
Formally, the global representation is computed as:
\begin{equation}
    P_g = \text{CrossAttn}(Q(P), K(h_g^S), V(h_g^S)) 
    \label{eq:crossattn}
\end{equation}
where $Q$, $K$, and $V$ 
are learnable weight matrices that project the input features into the query, key, and value subspaces, respectively.
The global representation $P_g$, summed with $P$, serves as the multi-level representation:
\begin{equation}
\label{eq:multi-level}
P_{\text{m-l}} = P + P_g,
\end{equation}
effectively encoding both local and global information for diffusion guidance.

\subsection{Diffusion Pair Guidance Stage}
\label{sec:method:diffusion-guidance}
Integrating cues from multiple pairs is non-trivial, as it demands careful interaction across signals without unintended cross-contamination. Multi-level global guidance further exacerbates the problem as global guidance should coordinate items coherence. Prior methods~\citep{mou2024t2i, zhang2023adding} typically pool all guidance information into a single aggregated representation, which can lead to mutual interference between pairs. Instead, we delegate the fusion to the pre-trained diffusion process itself: conditioning pairs are incorporated gradually across successive denoising iterations, allowing the model to assimilate them incrementally rather than through a single-step merging. Specifically, given $P_{\text{m-l}}$ we steer the diffusion process through cross-attention mechanisms and augment the frozen denoising network with an extra set of learnable cross-attention modules $\hat{w}$. Specifically, after each existing cross-attention layer $w$, we insert a parallel adapter that operates on the same feature input. These added modules incorporate the conditioning multi-level information $P_{\text{m-l}}$ at every diffusion step, enabling a progressive and iterative integration of information rather than a single-shot fusion. Formally, let $x$ be the input to a global text-conditioning cross-attention layer in the denoiser. The resulting conditioned features $x'$ produced by the paired attention blocks are given by:
\begin{equation}
x' = w(x, h^{T_g}) + \alpha \hat{w}(x, P_{\text{m-l}}),
\label{eq:diff-guidance}
\end{equation}
where $w(\cdot,\cdot)$ denotes standard cross-attention between two token sequences, and $h^{T_g}$ corresponds to the embedding of the global context description $T_g$, which conveys high-level semantic attributes such as style or background. The scalar $\alpha \in [0,1]$ controls the influence of the additional conditioning signal $P_\text{m-l}$. During training, we empirically set $\alpha=1$ so that the newly introduced attention adapters can fully learn how to combine the conditioning information.

Importantly, since these adapters are attention-based, they support a variable-length conditioning sequence. As a result, \methodshort can accommodate an arbitrary number of localized pairs without architectural changes or pooled representations.

%% file: sections/dataset.tex
\section{The \dataset dataset}
For model training and comparative evaluation, we introduce \dataset{}, a new dataset built on Fashionpedia~\citep{jia2020fashionpedia} for multi-localized conditional image generation.
In the following, we explain in detail how we organize the dataset based on garments and the localized text-sketch creation, as well as the dataset statistics.

\subsection{Local Garments Organization}\label{sec:dataset:garment_org}
We build \dataset{} on Fashionpedia~\cite{jia2020fashionpedia}, a dataset composed of 46k images for training and 1.2k for testing, where fashion experts annotate garments with fine-grained attributes and segmentation masks. While these masks include detailed part annotations (\eg, pockets, zippers, sleeves), they lack a hierarchical structure linking garment components. To improve compositionality, we introduce a two-level hierarchical organization based on segmentation mask overlaps.
Specifically, following Fashionpedia taxonomy, the 330k item annotations are first categorized into 14 ``whole-body items'', \ie, top-level garments such as tops, shirts, and skirts, regardless of the actual body coverage, and 32 ``garment parts'' (\eg, sleeves, pockets, and necklines). To ensure high-quality compositional annotations, we retain all whole-body categories, along with 21 sub-item categories from Fashionpedia, while removing 11 categories (31k annotations) that are rare or lack consistent overlap with any whole-body items, such as umbrellas, bags, and glasses. The statistics of the whole-body and garments parts are reported in Fig.~\ref{fig:dataset}.
Then, for each image, we determine the overlap between each garment part's mask and every whole-body item mask. 
Finally, we pre-process the images by resizing them to 512 pixels. We preserve their aspect ratio into a square format with white padding, to maintain visual consistency across samples. 

\subsection{Textual Annotation Creation}\label{sec:dataset:annotation}
Whole-body text annotations are considered as top-level annotations, \ie, a garment in the image, while part annotations are assigned to the whole-body item with which they have the greatest overlap, \ie, they are considered sub-garment annotations referring to an element, such as sleeves, necklines, and pockets. While the Fashionpedia annotations are rich in attribute set, they lack coherent natural language description, which is essential for the text conditioning. Thus, we generate the textual description for each garment in the image by prompting a pre-trained Large Language Model~\cite{touvron2023llama} with the hierarchical attributes of each garment, along with some in-context learning examples of the desired format.
Specifically, the LLM is instructed to act as a fashion expert, synthesizing the provided information (\ie, \textit{categories} and \textit{attributes} as in Fig.~\ref{fig:dataset}, top left) into a cohesive, opinion-free description (\ie, textual descriptions in Fig.~\ref{fig:dataset}, top right) that preserves the structural hierarchy of the items within a 90-token limit.
Notably, original Fashionpedia has no color information included, since its main focus is in explaining the fashion ontology of the different garments. 
This represents a notable gap, as color is a fundamental design component in sketch-based image generation. To address this limitation, we introduce color extraction as a dedicated task, extending the original conference paper~\cite{girella2025lots}, which includes attribute annotations from Fashionpedia but lacks color information.

To obtain color annotations, we rely on vision–language models to generate concise color descriptions aligned with human perception. We generate a garment-only white-background image for each garment instance using its segmentation mask. The white background minimizes background interference. To ensure small garments or accessories are sufficiently large to retain fine color details, we further inspect the image resolution by upsampling the image using Lanczos interpolation~\cite{lancoz} when the longest side of the garment mask is below 256 pixels.
As a single garment may contain several colors, we thus prompt a vision-language models (\texttt{SmolVLM-256M-Instruct}~\cite{marafioti2025smolvlm}) by quering ``\textit{What are the main colors? Ignore the white background.}'' to extract one to three color terms per garment instance. 
The resulting color descriptors are directly incorporated into the textual conditioning used during training. 
Fig.~\ref{fig:dataset} illustrates the hierarchical structure extracted from Fashionpedia alongside the corresponding global description generated by the LLM.
Finally, to account for potential hallucinations and errors from LLMs and VLMs, we manually inspected 50\% of the test samples and observed an acceptance rate of around 95\%, indicating high annotation reliability.

\subsection{Localized Sketches Creation}\label{sec:dataset:sketch}
One of the fundamental innovative aspect of the Sketchy dataset is the presence of localized garment-level sketches. 
These are primarily generated in an automated manner from the ground-truth images, using a pre-trained Image-to-Sketch model~\cite{li2019photo}. We remove background information via masking to ensure each sketch contains only information regarding the associated item, allowing for possible overlap in the cross-garment boundaries. 
Furthermore, we provide a global composition of all the garment sketches, which depicts the sketch of the entire outfit in the original image. The resulting sketches resemble high quality creations made by designers, as exemplified in Fig.~\ref{fig:sketch}, with natural contours and geometry.

\subsection{Sketchy in the Wild}\label{sec:dataset:sketch}
\begin{figure}
    \centering
    \includegraphics[width=\linewidth]{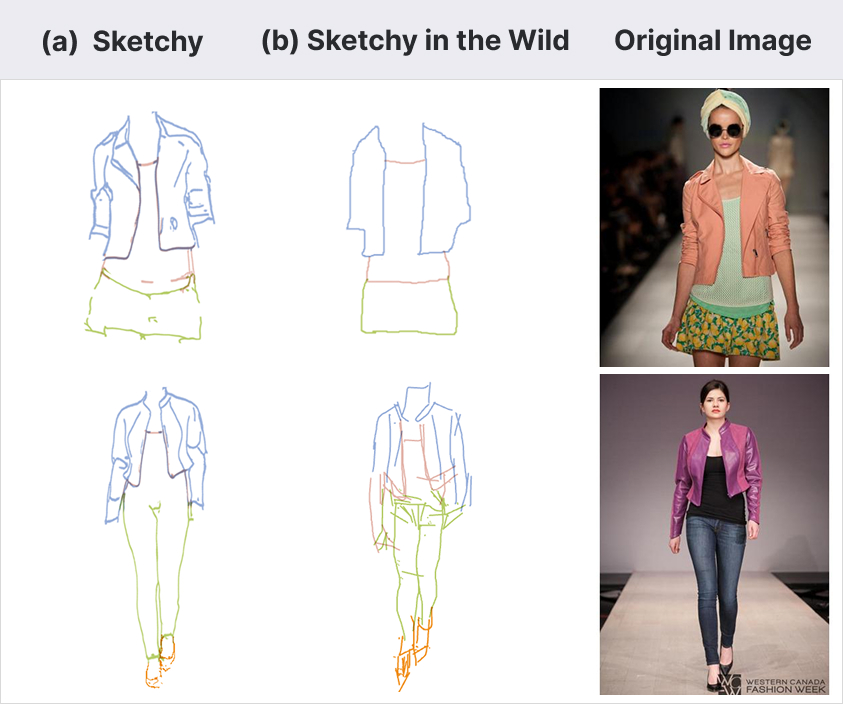}
    \caption{Examples of sketches in the \dataset and \dataset in the Wild dataset. The left column presents automatically annotated sketches in \dataset. The middle column shows collected human-drawn sketches in the \dataset in the Wild subset. The right column displays the corresponding original fashion images. Human-drawn sketches exhibit higher subjective abstraction and stylistic variability.} 
    \label{fig:sketch} 
\end{figure}

As a further contribution to \cite{girella2025lots}, we design an in-the-wild subset, made by non-expert people who draw with common tools, \eg either with the stylus, or mouses. This partition is named \emph{Sketchy in the Wild}, and is aimed at evaluating model robustness and generalization. In practice, Sketchy in the Wild consists of sketches derived from a uniform subsample of \dataset, retaining the original local and global captions while replacing the sketches with drawings created by a general audience.
To this sake, we developed a dedicated web-based annotation platform. Annotators draw on a standardized 512×512 white canvas using a multi-layer drawing interface, where each garment type is assigned to an independent layer.
This design allows annotators to sketch multiple garments within the same image while keeping strokes for different garments separated, enabling clear garment-level supervision.
During the human sketch collection, we display the original fashion image and the corresponding target garment region as visual references, in order to guide annotators to focus on the overall silhouette, key structural components, relative proportions, and salient pattern cues of each garment, rather than free-form artistic drawing. The platform supports both mouse-based and stylus-based input. Completed sketches are exported as images for model training, together with metadata such as input device, timestamps, and annotation status. The multi-layer design reduces annotator cognitive load and helps ensure consistent sketch quality across garments, while simplifying downstream dataset preparation. 
In total, we collected 141 sketches from 10 annotators aged between 20 and 45 years old, including 5 male and 5 female participants. The annotators came from diverse academic and professional backgrounds unrelated to art, fashion, or design. None had formal training in drawing or visual arts, and all self-reported only basic or occasional sketching experience. Participants used either a mouse (97) or a standard stylus (44) on consumer-grade devices, without access to professional illustration tools. See Fig.~\ref{fig:sketch} for a reference.
\begin{figure}
    \centering
    \includegraphics[width=\linewidth]{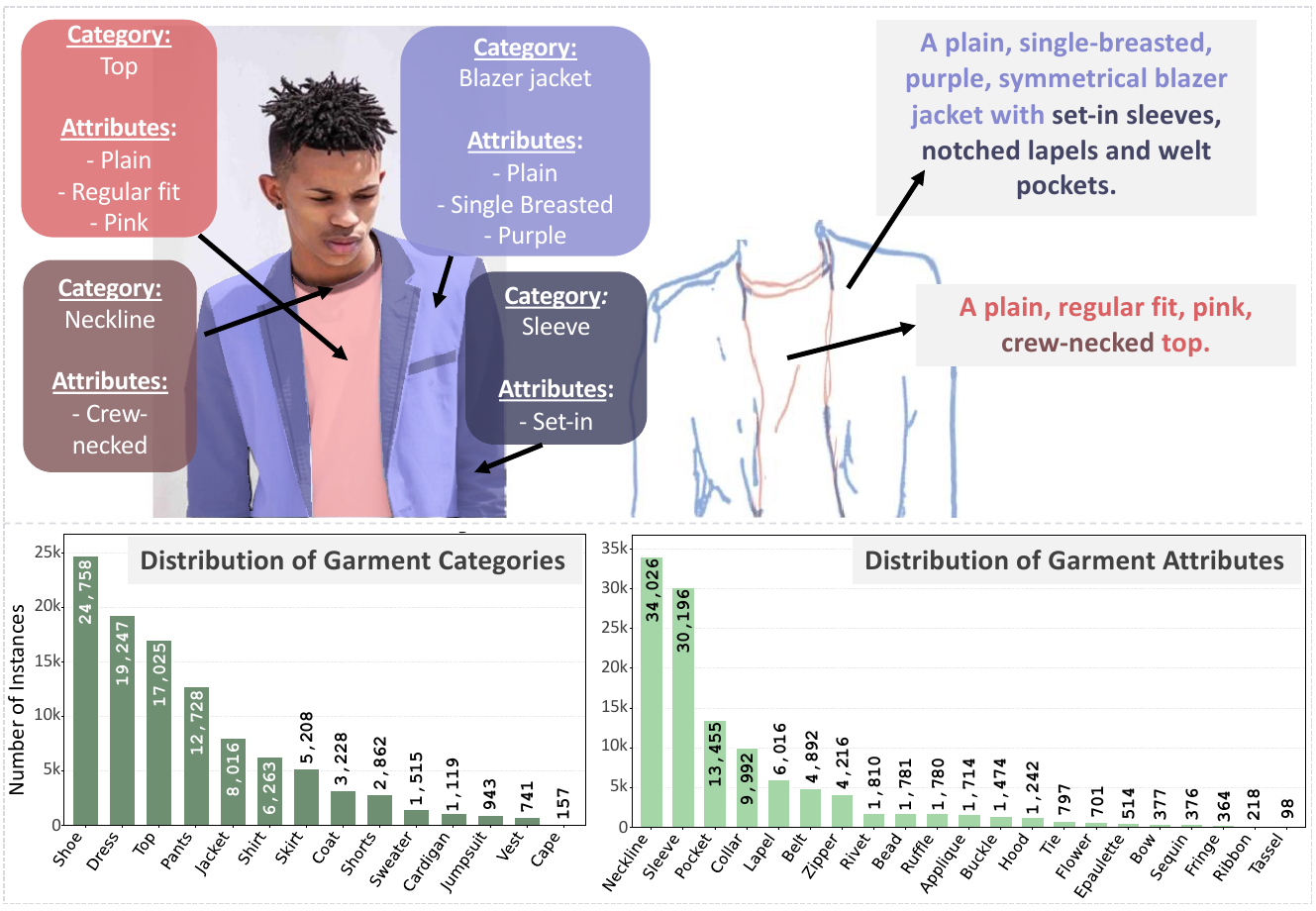}
    \caption{Overview of \dataset. We build a hierarchical structure by pairing the garment part annotations to their related whole-body garment. Then, garment-level sketches and natural language descriptions are added based on off-the-shelf models and the in-the-wild sketch collection pipeline. The bar charts illustrate the frequency of the prevalent categories within the dataset, representing the total count of annotations where whole-body items (left) and garment parts (right) appear.}
    \label{fig:dataset}
\end{figure}

\subsection{Dataset Statistics}\label{sec:dataset:statistics} Our \dataset extends Fashionpedia, providing a total of 47k images and 104k garment-level annotations, resulting in an average of 2.2 garment annotations per image (min 1, max 6). 
As shown in Fig.~\ref{fig:dataset}, each annotation contains the associated sketch, hierarchical attributes, and natural language description of the item. The average word length of the descriptions is 16 words. 
To examine annotation quality and potential structural bias of casual sketches from different input devices, we compute the Structural Similarity Index (SSIM) between the sketches
and the corresponding ground-truth images. 
Casual sketches achieve SSIM values comparable to synthetic sketches (±1\%), indicating a similarly strong structural alignment with the ground truth.
We further analyze the results by input device: stylus-drawn sketches attain an average SSIM roughly 0.015 higher in value than mouse-drawn sketches, with slightly greater variance.
The difference is not statistically significant, indicating that variations introduced by different input devices are stylistic in nature and do not result in fundamental differences in structural adherence.



%% file: sections/experiments.tex
\section{Experiments}\label{sec:experiments}
In this section, we present a comprehensive evaluation of \methodshort in comparison to state-of-the-art methods.
We begin by introducing the experimental setup (Sec.~\ref{subsec:metrics}), followed by main evaluation results, covering quantitative performance, generalization to casual sketches, qualitative results, and ablation studies (Sec.~\ref{subsec:comparisons}). Finally, we conduct human studies to further assess alignment with human preferences~(Sec.~\ref{subsec:human}).

Compared to the conference version~\cite{girella2025lots}, the experimental evaluation is extended along three new axes:  (i) localized semantic alignment measured at the garment level, (ii) analysis of global-local conditioning strategies enabled by the proposed multi-level architecture, and (iii) robustness to casual sketches.


\subsection{Experimental setup}\label{subsec:metrics}

\subsubsection{Compared Baselines}\label{subsec:baselines}
We compare against representative baselines and state-of-the-art approaches on text-to-image and sketch-to-image generation.
Regarding \textit{text-only diffusion models}, we compare against Stable Diffusion 1.5 (SD)~\cite{rombach2022high} and Stable Diffusion XL (SDXL)~\cite{podell2023sdxl}, which generate images solely from a global textual prompt without any explicit spatial or sketch guidance. We also compare with GLIGEN \cite{li2023gligen}, which enables localized textual conditioning through bounding-boxes.

Regarding \textit{sketch-to-image approaches}, we compare with methods that incorporate sketch guidance into pre-trained diffusion models, including ControlNet~\cite{zhang2023adding}, T2I-Adapter~\cite{mou2024t2i} and IP-Adapter~\citep{ye2023ip} both in a zero-shot manner and with fine-tuning on our \dataset dataset. 
All of these methods condition generation on a global sketch together with a global text prompt.
To further examine their ability to handle compositional inputs, we adapt ControlNet and T2I-Adapter to accept multiple local sketches while still relying on a single global textual description, denoted as Multi-ControlNet and Multi-T2I-Adapter, respectively.

Finally, we evaluate AnyControl~\cite{sun2024anycontrol}, a recent \textit{unified multi-control} approach supporting localized sketch conditioning with a global textual prompt.
We additionally include LOTS*, our prior conference version~\cite{girella2025lots}, as a strong multi-control baseline, where the asterisk distinguishes it from \methodshort, the method introduced in this extension.

\subsubsection{Evaluation Settings}
We consider two evaluation settings: (i) \emph{in-domain} performance, and (ii) \emph{generalization to casual sketches}. 
Under the in-domain setting, all models are trained and evaluated on the \dataset dataset, while under the generalization setting, models are trained on \dataset and evaluated on \emph{\dataset in the Wild}.

To ensure a fair comparison across models, conditioning inputs is adjusted to each generative model according to its specific input requirements. 
Specifically, for models that are designed to only take a single textual prompt, such as SD and SDXL, we concatenate all garment descriptions into a single global   description fed as model input.
For models requiring a single global sketch as the conditioning input, such as ControlNet, IP-Adapter and T2I-Adapter, we construct a composite sketch by combining all individual garment sketches. For models that support localized control, including Multi-ControlNet, Multi-T2I-Adapter and AnyControl, we provide garment-specific sketches and/or corresponding garment descriptions as conditioning inputs. For fair comparison, the global context prompt of \dataset is fixed across all samples as ``\textit{A picture of a model posing, high-quality, 4k}''. For experiments exploring global context prompt variations, we set distinct global context descriptions in different experimental settings, as in Fig.~\ref{fig:qualitatives_for_qual} and Sec.~\ref{subsec:qual}.
\input{tables/table1}
\subsubsection{Performance Metrics} 
We report two groups of metrics to quantify the \textit{global quality} and the \textit{compositional alignment}. 
As a measure of global visual quality, we adopt the Fréchet Inception Distance (FID)~\cite{heusel2017gans}, following prior work~\cite{sun2024anycontrol,goel2024pair,lukovnikov2024layout,cheng2023adaptively,bashkirova2023masksketch}. FID evaluates visual fidelity at the distribution level by comparing the statistics of generated images set against those of the ground-truth images, rather than assessing individual samples. Lower FID values ($\downarrow$) indicate higher perceptual realism and closer alignment between the generated and real image distributions. In simple terms, FID answers the question ``Do the generated images, considered as a set, match the real image distribution in terms of overall visual realism and diversity?''
To assess global semantic alignment, we adopt the GlobalCLIP score~\citep{radford2021learning}, computed as the cosine similarity between the CLIP encodings of the generated image and ground-truth image, in line with~\cite{bashkirova2023masksketch}. Unlike distribution-level metrics such as FID, GlobalCLIP operates at the individual image level.
Higher GlobalCLIP scores ($\uparrow$) indicate stronger semantic correspondence and better adherence globally. In practice, it answers the question ``Does the generated image, as a whole, semantically match the ground-truth?''

To assess the localized compositional alignment, we first report the LocalCLIP score, building on~\cite{lukovnikov2024layout,kim2023dense}, to assess the visual alignment between local parts of the generated image and the ground-truth image. 
Specifically, we compute the cosine similarity between the CLIP embeddings~\citep{radford2021learning} of the masked garment regions of the generated image and of the ground-truth image.
Higher LocalCLIP scores ($\uparrow$) indicate an improved localized semantic alignment with the ground-truth. In practice, LocalCLIP metrics answers the question ``Do local regions of the generated image semantically match the corresponding regions of the ground-truth image?''
Moreover, to evaluate the semantic alignment between the generated images with textual descriptions, we report the VQAScore~\cite{lin2024evaluating}, a compositional semantic alignment metric that leverages Visual Question Answering models to query the presence and correctness of specified attributes. A larger VQAScore ($\uparrow$) suggests greater compositional alignment to the provided textual prompt. It answers the question ``Are the attributes mentioned in the text present in the image?''

Extending the localized evaluation protocol introduced in the prior work~\cite{girella2025lots}, we further assess attribute alignment at the garment level, taking into account structural hints as in~\cite{liu2025evaluating}. 
Specifically, we report the Localized-VQAScore (L-VQAScore): differently from VQAScore, which operates at the image level, L-VQAScore measures whether queried attributes are correctly associated with their intended garment regions. 
Concretely, attribute-specific questions are constructed for each garment instance, and a VQA model is queried on the corresponding localized image crops. The final L-VQAScore is obtained by averaging scores across all garment instances. This metric directly captures localization accuracy. The higher L-VQAScore values ($\uparrow$) indicate more precise and structurally consistent semantic grounding. In practice, it answers the question ``Are the attributes present on the correct garment, in the correct region?''

Finally, we assess the sketch-following capability by reporting the Structural Similarity Index Measure (SSIM)~\cite{goel2024pair,wang2024lineart}. Specifically, we compute SSIM between the generated images and their corresponding ground-truth images.
SSIM measures the preservation of structural patterns. Higher SSIM values ($\uparrow$) indicate a stronger structural alignment, \ie local sketches are better composed and followed. It answers the question ``Does the generated image preserve the intended structural patterns?'' 

\input{tables/table2}
\subsubsection{Implementation Details}
We adopt \texttt{DINOv2 vits14}~\cite{oquab2023dinov2} as sketch encoder. For the text encoder, following the findings in~\cite{podell2023sdxl}, we combine \texttt{OpenCLIP ViT-bigG} \cite{openclip} and \texttt{CLIP ViT-L} \cite{radford2021learning} by concatenating the penultimate text encoder outputs along the channel-axis. 

During training, all backbone parameters are kept fixed, and optimization is restricted to the image and text projection layers, the proposed Pair-Former module, and the newly introduced cross-attention blocks. All fine-tuned methods are trained on the training split of the \dataset dataset.
We train \methodshort following the standard Stable Diffusion training protocol~\cite{rombach2022high}, while all other fine-tuned methods are trained with their respective official implementations and default configurations. 
For \methodshort, we use the Adam~\cite{kingma2014adam} optimizer, a learning rate of 1e-5, and a total batch size of 32.
All images are generated at a resolution of \texttt{(512x512)} using each model’s default inference configuration.

\subsection{Main Experimental Results}\label{subsec:comparisons}

\subsubsection{Quantitative Results on \dataset} 
\begin{figure*}
    \centering
    \includegraphics[width=0.95\linewidth]{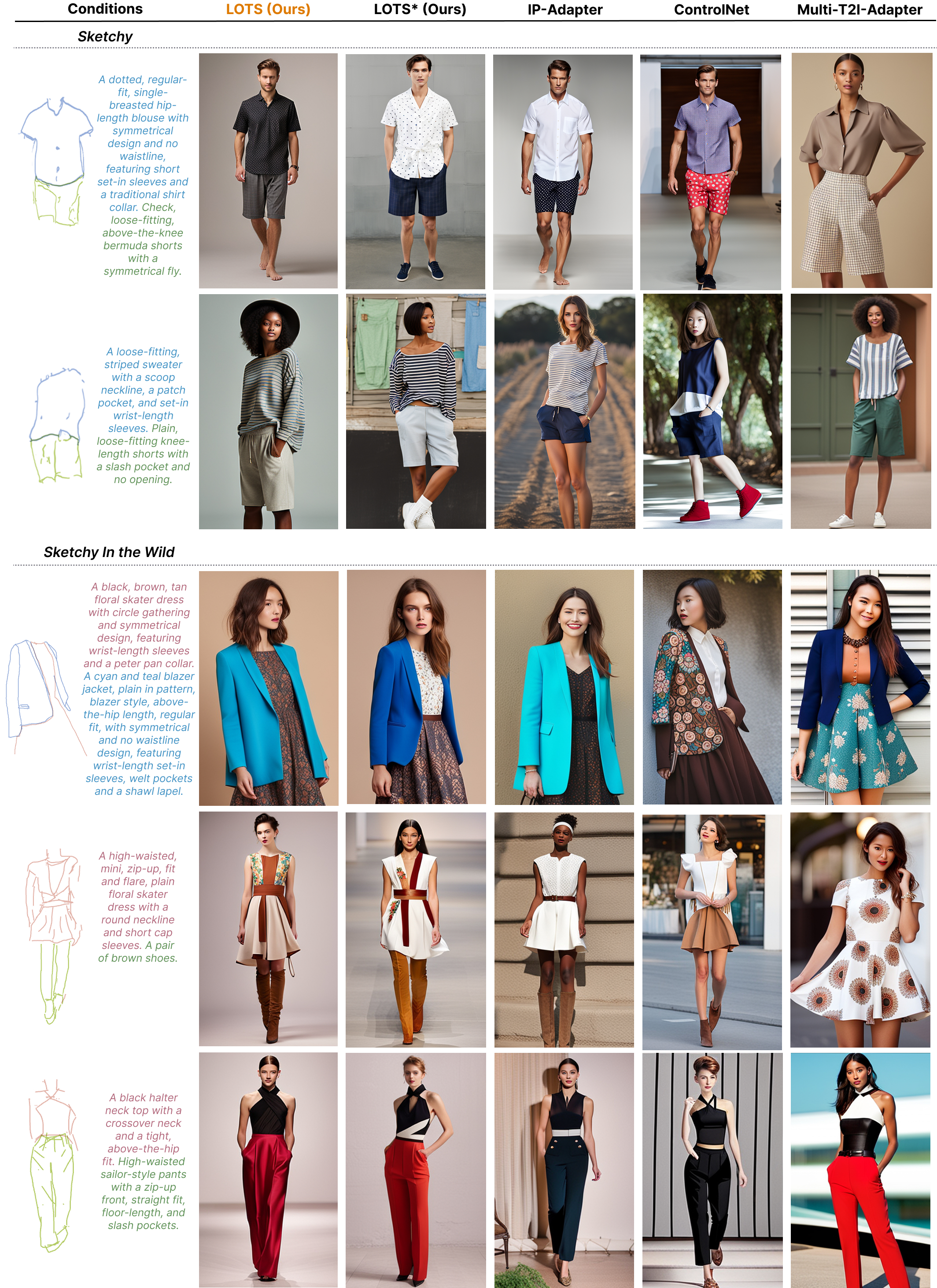}
    \caption{Qualitative comparison of \methodshort{} with our prior work LOTS*~\cite{girella2025lots}, ControlNet~\cite{zhang2023adding}, IP-Adapter~\cite{ye2023ip}, and Multi-T2I-adapter~\cite{mou2024t2i}, all in their fine-tuned versions. 
    Given localized sketch-text pairs as conditioning inputs, \methodshort{} better capture fine-grained attributes within the intended local regions of the generated images, effectively mitigating attribute confusion while maintaining strong global structural alignment.} 
    \label{fig:qualitatives}
\end{figure*}

Tab.~\ref{tab:vlm_stats} reports the performance of all compared methods spanning both global generation quality and compositional alignment on the test split of \dataset. The Table should be read together with Fig.~\ref{fig:qualitatives} (first three rows), to appreciate the quantitative results with some visual examples.

Notably, \methodshort scores the best performance on FID, GlobalCLIP, LocalCLIP and L-VQAScore, while ranking second in SSIM, demonstrating overall superiority and strong alignment both semantically and structurally.
Specifically, in terms of GlobalCLIP score, \methodshort surpasses all baselines, indicating strong semantic alignment at global levels. Fig.~\ref{fig:qualitatives} clearly highlights the performance gap between \methodshort and the weakest fine-tuned approach, Multi-T2I-Adapter (.561), which exhibits mismatches in garment type (\eg, sweater \textit{vs} t-shirt), color (\eg, black-brown \textit{vs} orange-teal), and texture (\eg, plain \textit{vs} dotted), producing images that, as a whole, do not semantically match the ground truth.

In terms of local alignment, the top performance of \methodshort in LocalCLIP (.826) and L-VQAScore (.700), indicates that local attributes are correctly placed in the generated images. Notably, \methodshort does not achieve the highest VQAScore (.706), which is instead obtained by ControlNet~\cite{zhang2023adding}. However, VQAScore only evaluates whether the attributes mentioned in the textual descriptions are present in the image, without verifying their correct spatial assignment. As a result, it is not well suited to detect attribute confusion. For instance, in the first row of Fig.~\ref{fig:qualitatives}, ControlNet correctly renders all attributes specified in the text, but assigns the dotted pattern to the trousers rather than to the shirt, thus exhibiting attribute confusion. These observations suggest that VQAScore alone is not a reliable metric for assessing localized attribute correctness. 


While some methods achieve higher performance on individual metrics (\eg zero-shot SDXL on VQAScore or IP-Adapter on SSIM), their improvements come with trade-offs in other metrics. Specifically, SDXL and the fine-tuned ControlNet achieve a higher VQAScore but sacrifice both global quality and sketch following, as evidenced by its significantly lower global quality and structural similarity metrics.
Although the zero-shot IP-Adapter attains the highest SSIM for structural alignment, it overemphasizes sketch guidance at the expense of prompt adherence and semantic alignment, which is evidenced by its low GlobalCLIP, LocalCLIP, VQAScore, and FID scores.
In conclusion, these results highlight that \methodshort effectively balances visual quality, structural adherence, and semantic alignment by jointly leveraging global and localized guidance, setting a new state-of-the-art for localized fashion sketch-to-image generation.

\subsubsection{Generalization to \dataset in the Wild} 
We further report results on Sketchy in the Wild, which contains human-collected casual sketches. In this setting, models are trained on the standard \dataset split and evaluated on the \dataset in the Wild subset.
It is  worth mentioning that this setting introduces a substantial domain gap due to the high variability in stroke style, abstraction, and structural distortion by individuals.

\begin{figure*}
    \centering
    \includegraphics[width=1.0\linewidth]{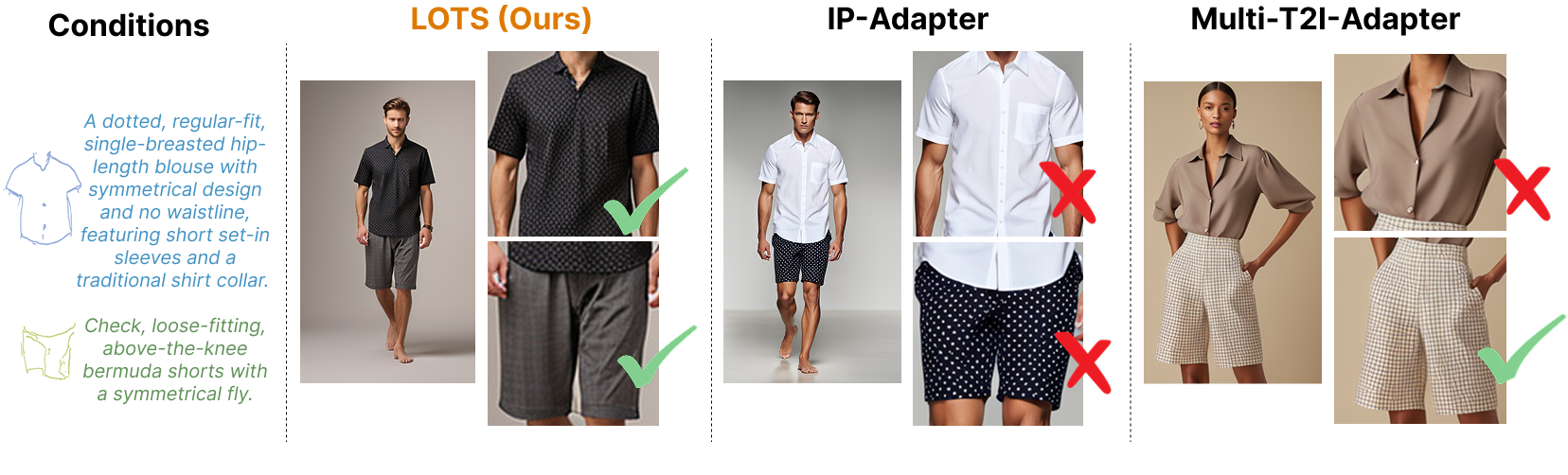}
    \caption{Qualitative comparison of \methodshort{} with IP-Adapter~\cite{ye2023ip}, and Multi-T2I-adapter~\cite{mou2024t2i}, all in their fine-tuned versions. Cropped views illustrate the details.} 
    \label{fig:qualitatives_attr}
\end{figure*}
\begin{figure}
     \centering
     \includegraphics[width=0.97\linewidth]{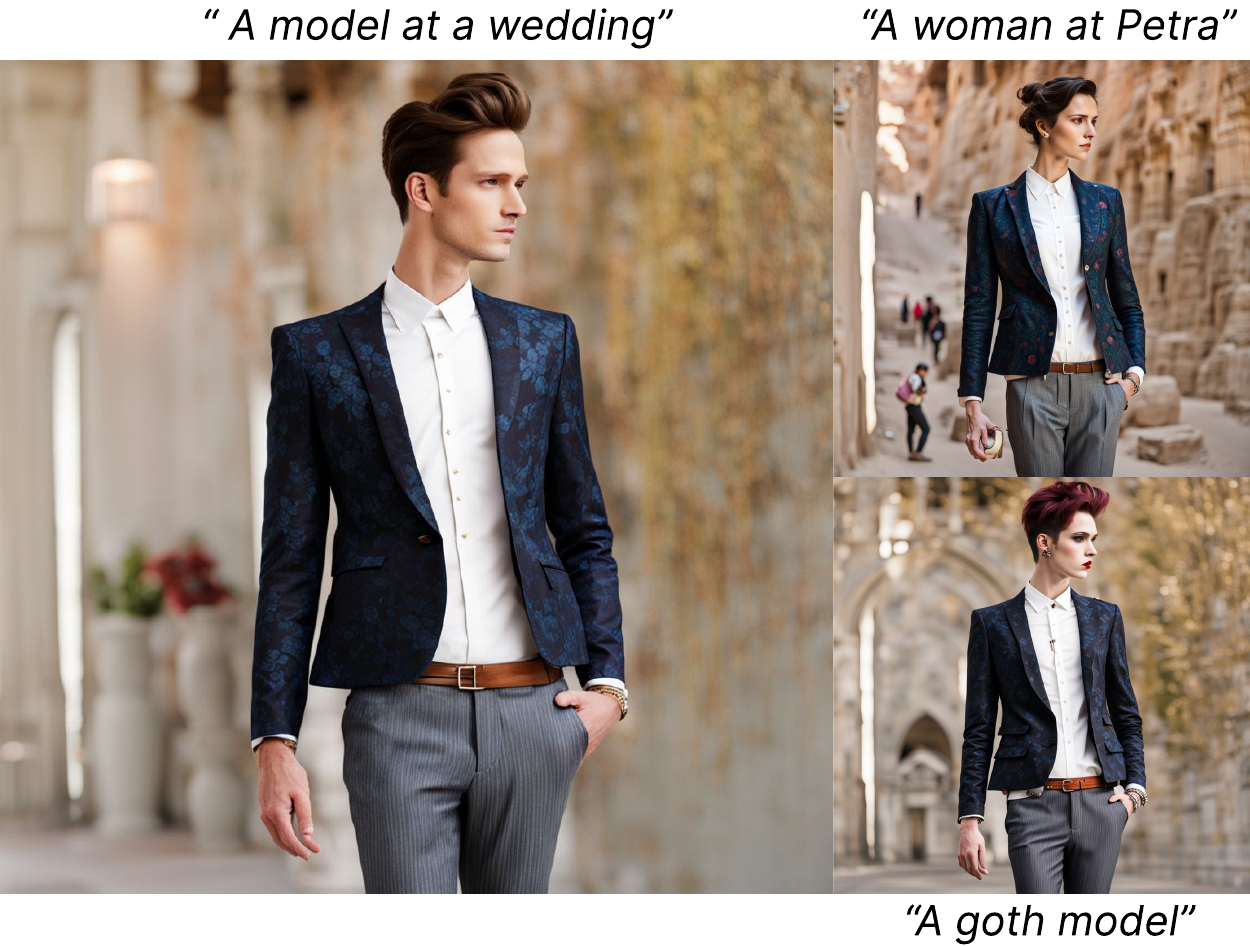}
     \caption{Effects of different global context descriptions on the generation of \methodshort. The text labels (top and bottom) indicate the specific global context prompts used in each generation. By changing it, we are able to customize general aspects such as the background and style of the model and the outfit.}
     \label{fig:qualitatives_for_qual}
\end{figure}
As visible in Tab.~\ref{tab:vlm_stats_human}, under such challenging scenario, performance degradation is observed across all methods.
Nevertheless, \methodshort consistently remains the best-performing method across most metrics, achieving strong semantic alignment and sketch adherence while demonstrating robustness to domain shift and variability in human-drawn sketches. Importantly, the relative ranking of \methodshort with respect to competing methods remains largely unchanged compared to the standard \dataset\ setting. Across most metrics, it maintains the same leading position. The only exception is the FID score (1.23), where it ranks third, following IP-Adapter~\citep{ye2023ip} (1.16) and LOTS*~\cite{girella2025lots} (1.19). This outcome is likely due to our method’s stronger adherence to sketch structure, which, when applied to the more noisy casual sketches in \dataset\ in the Wild, produces images that deviate more from the original ground-truth distribution.

\subsubsection{Qualitative Results}\label{subsec:qual}
Consider the example illustrated in Fig.~\ref{fig:qualitatives_attr}, the input description specifies two main items: ``\textit{a dotted blouse with traditional shirt collar, a check above-the-knee bermuda shorts}''. 
An accurate generation should place the ``\textit{dotted}'' pattern on the blouse while correctly rendering the shorts with the ``\textit{check}'' pattern. \methodshort can accurately reflect the intended patterns and garments. In contrast, compared top-performing baselines frequently fail to maintain correct attribute placement.
For example, in this case, IP-Adapter and ControlNet exhibit errors by placing the ``\textit{dotted}'' pattern on the shorts instead of the blouse, while Multi-T2I-Adapter fail to reflect it in the required blouse. 
Overall, ControlNet follows the sketch outline, yet it frequently omit or leak attributes, as shown in Fig.~\ref{fig:qualitatives}. For instance, when prompted to generate a ``\textit{floral skater dress}'' (row 3), it fails to render the floral pattern in the dress misassigning it to the blazer. This observation aligns with our human evaluation results, detailed in Sec.~\ref{subsec:human}: while ControlNet achieves high preference rate in the sketch alignment study, it demonstrates significantly lower performance in the attribute localization study.
Although other methods, such as Multi-T2I-Adapter, generate the attributes correctly in some examples, they show limited adherence to the sketches.
In contrast, \methodshort preserves semantic alignment and avoid attribute confusion while following the structural guidance provided by the sketch. 
These trends are consistently observed across multiple examples, 
as evidenced by the quantitative results
in Tab.\ref{tab:vlm_stats}, \ref{tab:vlm_stats_human}, and \ref{tab:attr_study}.

To illustrate the impact of our global context description $T_g$, we present in Fig.~\ref{fig:qualitatives_for_qual} three examples in which the local sketch and text are kept fixed, while the global context description varies. Notably, by varying the global context prompt, we manipulate the environmental context (\eg, shifting from a wedding venue to the ruins of Petra) and aesthetic style (\eg, a goth interpretation). Specifically, the description ``\textit{A goth model}'' results in an image with extra stylistic details, such as additional bracelets, earrings, pale skin, and red-tinted hair, while maintaining the overall properties of the garments.

\subsubsection{Ablation Study}

To understand the contribution of our design choices, we conduct an ablation study comparing different configurations: 
(i) \textit{LOTS*}, which utilize solely localized sketch-text pairs without explicit global sketch guidance, representing our preliminary conference version~\cite{girella2025lots}; 
(ii) \textit{CONCAT}, which incorporates global sketch information by directly concatenating global sketch features with local pair features; 
(iii) \textit{ATTN}, which integrates global sketch and local paired guidance through attention-based fusion;
(iv) \textit{64-TOKEN}, where the number of tokens in the Pair-Former pooling is increased from 32 to 64;
 (v) \textit{POOL} where pairs are merged via pooling and their integration is not deferred to the diffusion process; and (vi) our full model \textit{\methodshort}, which integrates global sketch and local paired guidance through attention-based fusion with 32-token pooling and merges the different pairs throughout the iterative denoising.
In particular, ATTN uses global sketch features as queries and localized paired features as keys and values when getting $P_g$, whereas in \methodshort, this is reversed (Eq.~\ref{eq:crossattn}). All other components and design choices are shared between the two. Similarly, 64-TOKEN adopts identical architectural design as \methodshort, differing only in the number of tokens used for the Pair-former pooling. Except 64-TOKEN, other variants rely 32 tokens.
\input{tables/ablation}
Table~\ref{tab:ablation} reports the results on the test split of \dataset in terms of global quality, semantic alignment, structural adherence, and compositional correctness. 
Notably, compared to our previous approach LOTS* as in~\cite{girella2025lots}, the new \methodshort achieves consistent improvements across most metrics, with around 6\% on FID and SSIM, and roughly 1\% on GlobalCLIP, LocalCLIP, and L-VQAScore, while maintaining comparable performance in VQAScore, with only a slight decrease of 0.49\%.
This indicates that the global conditioning design is important for perceptual quality and structural alignment.
With respect to the multi-level strategy,
simply concatenating global features, as in CONCAT, improves VQA-based metrics but degrades SSIM significantly, showing that naive global feature fusion emphasizes textual prompt following but tends to compromise structural alignment. 
In contrast, \methodshort effectively leverages global information via attention-based  fusion, achieving the best structural alignment and localized semantic alignment while maintaining competitive global image quality.
Furthermore, while ATTN achieves improvement upon LOTS* in compositional alignment metrics,
it is consistently outperformed by \methodshort. This demonstrates the effectiveness of our specific attention design, which leverages paired features as queries, improving global structural alignment without compromising fine-grained local semantics.
Interestingly, while the 64-token variant achieves comparable synthesis quality, it underperforms in sketch adherence and local semantic alignment, despite its increased training overhead. Finally, the significant gap in VQAScore and L-VQAScore between the POOL variant and LOTS highlights the risk of early conditioning aggregation, underscoring why deferring integration to the diffusion process is critical to avoid attribute confusion.
Overall, these results demonstrate that the design of \methodshort achieves a strong and efficient balance between image quality, semantic alignment, and structural adherence.


\subsection{Human Evaluation}
\label{subsec:human}
To further validate our findings, we design two user studies to evaluate model performance under human judgment, in line with prior works~\cite{girella2025lots,zhao2024uni,liu2025evaluating,jayasumana2024rethinking,kawar2023imagic}. 
Our studies involved 21 participants, who provided informed consent, forming a balanced gender distribution with an age range between 20-50 years old and diverse demographic backgrounds. In total, we collected 1525 responses, with an inter-annotator agreement rate~\cite{lin2024evaluating} of 92.5\%. 

For both user studies, we evaluate a total of nine models, consisting of \methodshort and a representative subset of the closest competitors selected based on performance across multiple metrics as in Tab.~\ref{tab:vlm_stats} and~\ref{tab:vlm_stats_human}.
Specifically, we include SDXL~\cite{podell2023sdxl} as a text-only, zero-shot reference, together with ControlNet~\cite{zhang2023adding} in its fine-tuned variant, as both exhibit strong performance in terms of VQAScore among the considered baselines; IP-Adapter~\citep{ye2023ip} in both zero-shot and fine-tuned settings, which exhibit competitive results with respect to SSIM and FID; and fine-tuned T2I-Adapter~\cite{mou2024t2i} and Multi-T2I-Adapter~\cite{mou2024t2i}, which exhibit balanced performance across metrics. 
This selection ensures a comprehensive comparison across conditioning strategies and performance profiles. 
\subsubsection{User Study Design}
The user studies aim to assess controllability rather than overall visual quality, with a focus on localized semantic alignment and structural adherence.

\noindent\textbf{\textit{Study I: attribute localization and leakage}}. In the study, images are generated from the considered models using the same sketch and textual descriptions, and then refined using the Stable Diffusion XL Refiner~\cite{sdxl-refiner} to avoid bias from the overall image quality. To better measure attribute localization, selected garment descriptions are enriched with a pattern (\eg, striped, dotted) and we ensure that the attribute appears only once in the target outfit to avoid ambiguity. Participants are asked to determine whether an attribute associated with the $i$-th garment is correctly localized in the intended garment and whether it incorrectly leaks to other garments. Based on these responses, we quantitatively evaluate the considered models regarding \textit{Precision ($\uparrow$)}, \textit{Recall ($\uparrow$)}, and \textit{F1 Score ($\uparrow$)} with respect to localized conditioning. Recall is defined as the fraction of times that a specified attribute is correctly applied to the intended clothing garment, while Precision measures how often the generated attribute appears exclusively on the intended item without being mistakenly applied to other objects. In particular, the F1 Score reflects the model’s overall effectiveness in balancing correct placement and reduced attribute confusion. Higher values indicate a stronger ability to localize attributes accurately while minimizing unintended attribute leakage. 
    
\noindent\textbf{\textit{Study II: human sketch adherence and structural alignment}}. Given a casual input sketch, participants are randomly presented with two images generated by different models from the same input in a side-by-side comparison. They are asked to select the image that better follows the provided sketch in terms of structural alignment, with the option to indicate a tie when both images are perceived as equally consistent. Model performance is quantified using the \textit{Preference Rate}, computed as the proportion of times a model is favored over \methodshort in pairwise comparisons. 


\input{tables/table3}
\subsubsection{Results with Human Evaluation}
Tab.~\ref{tab:attr_study} reports the results of the user studies.
In \textbf{\textit{Study I}}, \methodshort{} achieves the highest F1 and Precision scores across all models,
indicating that our method is highly effective at correctly assigning attributes to the intended garments while minimizing unintended leakage.
In \textit{\textbf{Study II}}, participants consistently preferred the images generated by \methodshort{} and ControlNet with respect to sketch following, indicating their superior ability to preserve the spatial proportions and layout specified by the input sketches. 
While ControlNet achieves the highest sketch adherence scores, its improvement comes at the cost of weaker semantic alignment and substantially increased attribute confusion (\eg, Fig.~\ref{fig:qualitatives} rows 1 and 3) and reduced realism (\eg, Fig.~\ref{fig:qualitatives} last two rows). 
\methodshort{} achieves significantly better semantic alignment while maintaining comparable sketch adherence.
These results collectively show that \methodshort{} achieves high attribute localization accuracy and stronger sketch adherence, effectively balancing semantic and structural alignment.


%% file: tables/table1.tex
\begin{table*}
\caption{Comparisons between \methodshort and state-of-the-art sketch-to-image approaches on \dataset. In the Conditioning column, L and G indicate whether the model accepts Local or Global inputs as Visual or Textual conditioning. We divide the table into three sections: \inlineColorbox{zshot-blue}{zero-shot} approaches, \inlineColorbox{ft-red}{fine-tuned} approaches, 
our prior and current proposed method \inlineColorbox{method-green}{LOTS*} and 
\inlineColorbox{method-green}{\methodshort}. We highlight the best performance in \textbf{bold} and underline the \underline{second best}.}\label{tab:vlm_stats}
    \centering
    \resizebox{\linewidth}{!}{
    \begin{tabular}{l c | ccc cccc}
        \toprule
        \multirow{2}{*}{\textbf{Model}} & \multirow{2}{*}{\textbf{Conditioning}}& \multicolumn{2}{c}{\textbf{Global Quality}} 
        &\multicolumn{4}{c}{\textbf{Compositional Alignment}}\\ 
        \cmidrule(lr){3-4}
        \cmidrule(lr){5-8}
         & Visual/Textual & FID ($\downarrow$) & GlobalCLIP ($\uparrow$) & LocalCLIP ($\uparrow$) & VQAScore ($\uparrow$) & L-VQAScore ($\uparrow$) & SSIM ($\uparrow$)\\
        \midrule
        \cellcolor{zshot-blue}{SD~\cite{rombach2022high}} & -/G & 1.03 & .613 & .753 & .694 & .430 & .652\\
        \cellcolor{zshot-blue}{SDXL~\cite{podell2023sdxl}} & -/G & 1.09 & .567 & .757 & \textbf{.781} & .547 & .661\\
        \cellcolor{zshot-blue}{GLIGEN~\cite{li2023gligen}} & -/L & 1.10 & .544 & .709 & .282 & .213 & .594\\
        \cellcolor{zshot-blue}{ControlNet~\cite{zhang2023adding}} & G/G & 0.96 & .633 & .801 & .709 & .579 & .623\\
        \cellcolor{zshot-blue}{Multi-ControlNet~\cite{zhang2023adding}} & L/G & 0.98 & .624 & .786 & .687 & .528 & .638\\
        \cellcolor{zshot-blue}{IP-Adapter~\cite{ye2023ip}} & G/G & 2.48 & .540 & .706 & .432 & .326 & \textbf{.710}\\
        \cellcolor{zshot-blue}{T2I-Adapter~\cite{mou2024t2i}} & G/G& 2.03 & .543 & .726 & .644 & .506 & .503\\
        \cellcolor{zshot-blue}{Multi-T2I-Adapter~\cite{mou2024t2i}} & L/G& 2.25 & .520 & .709 & .527 & .371 & .489\\
        \cellcolor{zshot-blue}{AnyControl~\citep{sun2024anycontrol}} & L/G & 1.06 & .608 & .788 & .688 & .554 & .495\\
        \midrule
        \cellcolor{ft-red}{GLIGEN~\cite{li2023gligen}} & -/L & 1.12 & .570 & .734 & .330 & .292 & .511\\
        \cellcolor{ft-red}{ControlNet~\cite{zhang2023adding}} & G/G & 0.82 & \underline{.655} & .812 & \underline{.760} & .652 & .583\\
        \cellcolor{ft-red}{Multi-ControlNet~\cite{zhang2023adding}} & L/G & 0.96 & .634 & .799 & .722 & .610 & .546\\
        \cellcolor{ft-red}{IP-Adapter~\cite{ye2023ip}} & G/G & \underline{0.75} & .621 & .799 & .751 & .590  & .637\\
        \cellcolor{ft-red}{T2I-Adapter~\cite{mou2024t2i}} & G/G & 1.25 & .571 & .753 & .726 & .536 & .597\\
        \cellcolor{ft-red}{Multi-T2I-Adapter~\cite{mou2024t2i}} & L/G & 1.36 & .561 & .741 & .754 & .487 & .592\\
        \cellcolor{method-green}LOTS* \textbf{(Ours)}~\cite{girella2025lots}  & L/L & 0.79 & .651 & \underline{.818} & .709 & \underline{.692} & .651\\
        \cellcolor{method-green}\methodshort  \textbf{(Ours)}& L/L & \textbf{0.74} & \textbf{.660} & \textbf{.826} & .706 & \textbf{.700} & \underline{.691}\\ 
        \bottomrule
    \end{tabular}
}
\end{table*}

%% file: tables/table2.tex
\begin{table*}[t]
\caption{Comparisons between \methodshort and state-of-the-art sketch-to-image approaches on \dataset in the wild split. 
In the Conditioning column, L and G indicate whether the model accepts Local or Global inputs as Visual or Textual conditioning. We divide the table into three sections: \inlineColorbox{zshot-blue}{zero-shot} approaches, \inlineColorbox{ft-red}{fine-tuned} approaches, our 
prior and current proposed method \inlineColorbox{method-green}{LOTS*} and 
\inlineColorbox{method-green}{\methodshort}. We highlight the best performance in \textbf{bold} and underline the \underline{second best}.
}\label{tab:vlm_stats_human}
    \centering
    \resizebox{\linewidth}{!}{
    \begin{tabular}{l c | ccc cccc}
        \toprule
        \multirow{2}{*}{\textbf{Model}} & \multirow{2}{*}{\textbf{Conditioning}}& \multicolumn{2}{c}{\textbf{Global Quality}} 
        &\multicolumn{4}{c}{\textbf{Compositional Alignment}}\\ 
        \cmidrule(lr){3-4}
        \cmidrule(lr){5-8}
         & Visual/Textual & FID ($\downarrow$) & GlobalCLIP ($\uparrow$) & LocalCLIP ($\uparrow$) & VQAScore ($\uparrow$) & L-VQAScore ($\uparrow$) & SSIM ($\uparrow$)\\
        \midrule
        \cellcolor{zshot-blue}{SD~\cite{rombach2022high}} & -/G & 1.46 & .614 & .759 & .703 & .438 & .645\\
        \cellcolor{zshot-blue}{SDXL~\cite{podell2023sdxl}} & -/G & 1.46 & .566 & .759 & \textbf{.787} & .541 & .661\\
        \cellcolor{zshot-blue}{GLIGEN~\cite{li2023gligen}} & -/L & 1.57 & .538 & .709 & .293 & .223 & .599\\
        \cellcolor{zshot-blue}{ControlNet~\cite{zhang2023adding}} & G/G & 1.37 & .620 & .786 & .710 & .540 & .616\\
        \cellcolor{zshot-blue}{Multi-ControlNet~\cite{zhang2023adding}} & L/G & 1.37 & .619 & .782 & .683 & .513 & .635\\
        \cellcolor{zshot-blue}{IP-Adapter~\cite{ye2023ip}} & G/G & 2.75 & .539 & .708 & .406 & .319 & \textbf{.714}\\
        \cellcolor{zshot-blue}{T2I-Adapter~\cite{mou2024t2i}} & G/G& 2.54 & .535 & .725 & .638 & .522 & .495\\
        \cellcolor{zshot-blue}{Multi-T2I-Adapter~\cite{mou2024t2i}} & L/G& 2.64 & .521 & .707 & .495 & .365 & .487\\
        \cellcolor{zshot-blue}{AnyControl~\citep{sun2024anycontrol}} & L/G & 1.41 & .602 & .792 & .697 & .568 & .538\\
        \midrule
        \cellcolor{ft-red}{GLIGEN~\cite{li2023gligen}} & -/L & 1.58 & .566 & .732 & .322 & .317 & .512\\
        \cellcolor{ft-red}{ControlNet~\cite{zhang2023adding}} & G/G & 1.24 & \underline{.635} & .799 & \underline{.763} & .649 & .565\\
        \cellcolor{ft-red}{Multi-ControlNet~\cite{zhang2023adding}} & L/G & 1.37 & .626 & .793 & .734 & .604 & .535\\
        \cellcolor{ft-red}{IP-Adapter~\cite{ye2023ip}} & G/G & \textbf{1.16} & .603 & .788 & .757 & .601 & .673\\
        \cellcolor{ft-red}{T2I-Adapter~\cite{mou2024t2i}} & G/G & 1.60 & .582 & .752 & .757 & .511 & .606\\
        \cellcolor{ft-red}{Multi-T2I-Adapter~\cite{mou2024t2i}} & L/G & 1.72 & .570 & .739 & .741 & .481 & .599\\
        \cellcolor{method-green}LOTS* \textbf{(Ours)}~\cite{girella2025lots}  & L/L & \underline{1.19} & .629 & \underline{.808} & .749 & \underline{.663} & .673\\
        \cellcolor{method-green}\methodshort  \textbf{(Ours)}& L/L & 1.23 & \textbf{.636} & \textbf{.817} & .735 & \textbf{.684} & \underline{.709}\\ 
        \bottomrule
    \end{tabular}
}
\end{table*}

%% file: tables/ablation.tex
\begin{table*}[t!]
\caption{Ablation study comparing the impact of global conditioning, pair-former pooling and diffusion guidance on key metrics. We highlight the best performance in \textbf{bold}. }\label{tab:ablation}
\centering
\resizebox{1.0\textwidth}{!}{
\begin{tabular}{l c cc cccc}
\toprule
\multirow{2}{*}{\textbf{Component}} & \multirow{2}{*}{\textbf{Choice}}& \multicolumn{2}{c}{\textbf{Global Quality}} &\multicolumn{4}{c}{\textbf{Compositional Alignment}}\\ 
\cmidrule(lr){3-4}
\cmidrule(lr){5-8}
& & FID ($\downarrow$) & GlobalCLIP ($\uparrow$) & LocalCLIP ($\uparrow$) & VQAScore ($\uparrow$) & L-VQAScore ($\uparrow$) & SSIM ($\uparrow$)\\
\midrule
        \multirow{3}{*}{Global Conditioning}
         & LOTS* \textbf{(Ours)} & 0.79 & .651 & .818 & .709 & .692 & .651 \\
         &CONCAT & 0.79 & .651 & .817 & \textbf{.769} & \textbf{.703} & .644 \\ 
         &ATTN & 0.82 & .640 & .823 & .700 & .698 & .684 \\
        \midrule
        
        Pair-Former & 64-TOKEN & \textbf{0.70} & .656 & \textbf{.826} & .748 & .689 & .676 \\
        \midrule
        Diffusion Guidance & POOL & 0.74 & .654 & .823 & .680 & .612 & .687 \\
        \midrule
        & \methodshort \textbf{(Ours)} & 0.74 & \textbf{.660} & \textbf{.826} & .706 & .700 & \textbf{.691} \\
\bottomrule
\end{tabular}
}
\end{table*}

%% file: tables/table3.tex
\begin{table}[t]
\caption{Results of subjective user studies on attribute localization and structural alignment between \methodshort and selected top-performing models. For the attribute localization study, we highlight the best results in \textbf{bold} and underline the \underline{second best}.}\label{tab:attr_study}
    \centering
    \resizebox{0.48\textwidth}{!}{
    \begin{tabular}{l c c c c c}
        \toprule
        & \multicolumn{3}{c}{\textbf{Attribute Localization}} & \multicolumn{1}{c}{\textbf{Structural Alignment}} \\
        \cmidrule(lr){2-4} \cmidrule(lr){5-5}
        \textbf{Model} & Precision ($\uparrow$)& Recall ($\uparrow$) & F1 ($\uparrow$)  & Preference Rate \% ($\uparrow$) \\
        \midrule
        \cellcolor{zshot-blue}{SDXL~\cite{podell2023sdxl}} & .636 & \textbf{.754} & .690 & 14.7\\
        \cellcolor{zshot-blue}{IP-Adapter~\cite{ye2023ip}} & .625 & .139 & .227 & 0.00\\
        \cellcolor{ft-red}{ControlNet~\cite{zhang2023adding}} & .667 & .516 & .582 & 67.6\\
        \cellcolor{ft-red}{IP-Adapter~\cite{ye2023ip}} & .559 & .384 & .455 & 4.00\\
        \cellcolor{ft-red}{T2I-Adapter~\cite{mou2024t2i}} & .463 & .397 & .427 & 3.00\\
        \cellcolor{ft-red}{Multi-T2I-Adapter~\cite{mou2024t2i}} & .551 & \underline{.692} & .614 & 2.90\\
        \cellcolor{method-green}LOTS* \textbf{(Ours)}~\cite{girella2025lots} & \underline{.813} & .650 & \underline{.722} & 42.3\\
        \cellcolor{method-green}\methodshort  \textbf{(Ours)} & \textbf{.870} & .627 & \textbf{.729} & - \\
        \bottomrule
    \end{tabular}
    }
\end{table}

%% file: sections/conclusion.tex
\section{Conclusion}
In this work, we address the challenge of multi-localized sketch–text conditional image generation. We consider a realistic fashion design setting, where multiple garments must be synthesized coherently to preserve both global structure of the outfit and fine-grained semantics of garments. 
Extending our previous conference version, we propose LOTS, a multi-level conditioning framework that explicitly integrates localized sketch–text semantic pairs with global sketch guidance.
To support training and evaluation, we introduced Sketchy, a dataset that extends Fashionpedia with localized garment sketches, hierarchical textual descriptions and instance-level color annotations. We also introduce a new partition of Sketchy that contains casual sketches collected through a dedicated interactive platform. 
We conduct comprehensive experiments on both synthetic and human sketches, where LOTS achieves state-of-the-art performance in most quantitative metrics and human evaluations. LOTS effectively mitigates attribute confusion across local garments, while maintaining strong structural adherence and compositional consistency.

Future work will extend the framework to interactive and iterative design scenarios where users can progressively refine sketches and textual descriptions. Beyond fashion design, we will also explore novel applications of LOTS to other domains requiring fine-grained spatial and semantic control, such as interior design, industrial design, and character creation.


%% file: sections/ack.tex
\section{Acknowledgment}
This study was supported by LoCa AI, funded by Fondazione CariVerona (Bando Ricerca e Sviluppo 2022/23), PNRR FAIR - Future AI Research (PE00000013) and Italiadomani (PNRR, M4C2, Investimento 3.3), funded by NextGeneration EU.
This study was also carried out within the PNRR research activities of the consortium iNEST (Interconnected North-Est Innovation Ecosystem) funded by the European Union Next-GenerationEU (Piano Nazionale di Ripresa e Resilienza (PNRR) – Missione 4 Componente 2, Investimento 1.5 – D.D. 1058 23\/06\/2022, ECS\_00000043). This manuscript reflects only the Authors’ views and opinions. Neither the European Union nor the European Commission can be considered responsible for them.
We acknowledge ISCRA for awarding this project access to the LEONARDO supercomputer, owned by the EuroHPC Joint Undertaking, hosted by CINECA (Italy).
We acknowledge EuroHPC Joint Undertaking for awarding us access to MareNostrum5 as BSC, Spain.
Finally, we acknowledge HUMATICS, a SYS-DAT Group company, for their valuable contribution.